\documentclass[letterpaper]{article} %
\usepackage[submission]{aaai24}  %
\usepackage{times}  %
\usepackage{helvet}  %
\usepackage{courier}  %
\usepackage[hyphens]{url}  %
\usepackage{graphicx} %
\usepackage{natbib}  %
\usepackage{caption} %
\usepackage{amsfonts}       %
\usepackage{nicefrac}       %
\usepackage{microtype}      %
\usepackage{xcolor}
\usepackage{xspace}
\usepackage{subcaption}

\usepackage{booktabs}
\usepackage{pifont}

\usepackage[ruled,vlined,linesnumbered]{algorithm2e}
\newcommand{\SP}{${}$\hspace{0.25cm}}
\usepackage{amsmath}
\usepackage{amsfonts,amsthm}
\usepackage{amssymb}
\usepackage{bm}
\usepackage{cleveref}
\theoremstyle{plain}

\newtheorem{theorem}{Theorem}
\newtheorem{assumption}{Assumption}
\newtheorem{corollary}{Corollary}

\usepackage{mathtools}

\DeclareMathAlphabet{\mathcal}{OMS}{cmsy}{m}{n}

\usepackage{dsfont}

\usepackage{physics}
\usepackage{tipa}

\graphicspath{ {images/} }

\newcommand{\PseudoDefense}{\textit{Pseudo-Pruning}\xspace}
\newcommand{\algoname}{\textit{PriPrune}\xspace}

\title{\LARGE PriPrune: Quantifying and Preserving Privacy in Pruned Federated Learning}

\author{
    Tianyue Chu\equalcontrib \textsuperscript{1,}\textsuperscript{3},
    Mengwei Yang\equalcontrib \textsuperscript{2},
    Nikolaos Laoutaris\textsuperscript{1},
    Athina~Markopoulou\textsuperscript{2}
}
\affiliations{
    \textsuperscript{\rm 1}IMDEA Networks Institute\\
    \textsuperscript{\rm 2}University of California, Irvine\\
    \textsuperscript{\rm 3}Universidad Carlos III de Madrid\\
    \{tianyue.chu, nikolaos.laoutaris\}@imdea.org, 
    \{mengwey, athina\}@uci.edu,

}

\begin{document}

\maketitle

\begin{abstract} 
Federated learning (FL) is a paradigm that allows several client devices and a server to collaboratively train a global model, by exchanging only model updates, without the devices sharing their local training data. 
These devices are often constrained in terms of communication and computation resources, and can further benefit from model pruning -- a paradigm that is widely used to reduce the size and complexity of models. %
Intuitively, by making local models coarser, pruning is expected to also provide some protection against privacy attacks in the context of FL. However this protection has not been previously characterized, formally or experimentally, and it is unclear if it is sufficient against state-of-the-art attacks.  

In this paper, we perform the first investigation of privacy guarantees for model pruning in FL.
We derive information-theoretic upper bounds on the amount of information leaked by pruned FL models. We complement and validate these theoretical findings, with  comprehensive experiments that involve state-of-the-art privacy attacks, on several state-of-the-art FL pruning schemes, using benchmark datasets. This evaluation provides valuable insights into the choices and parameters that can affect the privacy protection provided by pruning.
Based on these insights, we introduce \algoname --  a privacy-aware algorithm for local model pruning, which uses a personalized per-client defense mask and adapts the defense pruning rate so as to jointly optimize privacy and model performance.  \algoname is universal in that can be applied after any pruned FL scheme on the client, without modification, and protects against any inversion attack by the server. 
Our empirical evaluation demonstrates that \algoname significantly improves the privacy-accuracy tradeoff compared to state-of-the-art pruned FL schemes that do not take privacy into account. 
For example, on the FEMNIST dataset, \algoname improves privacy by \textit{PruneFL} by 36.3\% without penalizing accuracy.

\end{abstract}

\section{Introduction}

Federated Learning (FL) has emerged as the predominant paradigm for distributed machine learning across a multitude of user devices ~\cite{mcmahan2017communication,kairouz2021advances}. It is known to have several benefits in terms of reducing the communication, computation and storage costs for the clients, training better global models, and raising the bar for privacy by not sharing the local data.
FL allows users to train models locally on their (client) devices without revealing their local data but instead collaborate by sharing only model updates that can be combined to build a global model through a central server.
A typical FL scenario involves numerous users and resource-constrained devices~\cite{li2020fedrated} 
making it challenging to perform resource-intensive tasks like training Deep Neural Networks (DNNs)  on them. 

Independently, significant efforts have been put toward optimizing sparse DNNs to create lightweight models suitable for training on edge devices~\cite{han2015learning,chen2020lottery,ma2021sanity}. 
Model pruning, which involves the removal of a certain percentage of parameters, has gained increasing attention and is being utilized in various ML tasks to reduce model complexity and induce sparsity~\cite{lecun1989optimal,han2015deep,park2023dynamic}.
The primary objective of model pruning is to derive a sparse ML model that decreases the computational requirements and enhances efficiency without compromising accuracy~\cite{frankle2019lottery}. %

Several model pruning techniques have been incorporated into the context of FL to further optimize model sparsity and meet the constraints of devices and communication bandwidth~\cite{jiang2022pruneFL,bibikar2022feddst,li2020lotteryfl}. The design of pruning for FL aims to strike a balance between model size and performance, enabling efficient and scalable FL applications under resource constraints.

Our key intuition is that using pruning methods in FL should improve privacy by virtue of their ability to reduce the model parameters. This reduction in parameters leads to a decrease in the amount of sensitive information stored in the model, thereby enhancing privacy protection against attacks that aim to use this information to reverse engineer the model to obtain the input data. %
Pruning methods typically focus solely on improving communication while preserving model accuracy,  without explicitly taking into account privacy, and the privacy impact of pruning methods has not been quantified before, theoretically or empirically. Our paper builds on top of current pruning methods and focuses on further pruning for privacy, optimizing the ACC-privacy tradeoff.
To the best of our knowledge, our study is the first one that evaluates the privacy benefit of model pruning within the context of FL, and further optimizes pruning for both inference and privacy.  We make two main contributions. 

First, {\bf we quantity privacy in model pruning.}
 We derive information-theoretic upper bounds on the amount of information revealed about any single user's dataset via model updates, in {\em any} pruned FL scheme,  to an honest-but-curious server. %
 We also conduct a comprehensive empirical evaluation that quantifies the exact amount of privacy leakage of six different pruning schemes, considering several state-of-the-art attacks, and employing several benchmark datasets. Within the Deep Leakage From Gradient (DLG) family of attacks~\cite{zhu2019deep}, we also design novel privacy attacks (which we refer to as Sparse Gradient Inversion -- SGI) specifically tailored to exploit vulnerabilities inherent to FL model pruning. %
This evaluation validates the theoretical analysis and also provides valuable insights into the choices and parameters that can affect the privacy protection provided by pruning.

Building upon these insights, we make our second contribution: we propose {\bf \algoname -- a privacy-preserving pruning mechanism.}
\algoname defends against DLG-type of privacy attack in pruned FL, by performing additional local pruning after any pruned FL scheme on the client, and before sending the update to the honest-but-curious server. \algoname applies a personalized per-client defense mask and adapts the defense pruning rat,e so as to jointly optimize model performance {\em and} privacy, using standard backpropagation augmented with Gumbel Softmax Sampling,
Another key idea incorporated into the design of \algoname is what we refer to as \PseudoDefense: the defense mask is only applied before sending a model update to the server, thus improving privacy; however, the parameters are not actually pruned, but retained for local training in next rounds, thus maintaining model accuracy.
We evaluate the efficacy of \algoname through comprehensive experiments across 6 FL pruning schemes, a state-of-the-art attack and several benchmark datasets, and we show that it achieves a significantly better privacy-accuracy trade-off than privacy-unaware pruned FL baselines. For example, on the FEMNIST dataset, \algoname improves privacy \textit{PruneFL}~\cite{jiang2022pruneFL} by 36.3\% without accuracy loss.

\section{Related Work}
\label{sec:related}
\paragraph{Neural Network Pruning}
Model pruning involves selecting a sparse sub-network from a larger neural network by removing a certain percentage of weights or connections while striving to maintain its accuracy at a high level. %
Early studies in model pruning have investigated a range of pruning criteria, encompassing methods such as first-order~\cite{mozer1988skeletonization,karnin1990simple,molchanov2016pruning} and second-order~\cite{lecun1989optimal,hassibi1992second,dong2017learning} Taylor expansions as well as more sophisticated variants~\cite{yu2018nisp, molchanov2019importance}. In recent times, magnitude-based pruning~\cite{han2015deep,han2015learning,park2020lookahead} has gained popularity as a prevailing approach, wherein network parameters with sufficiently small magnitudes are systematically removed from the neural network.
Other recent state-of-the-art studies have introduced various gradient-based one-shot pruning algorithms at initialization. \textit{Snip}~\cite{lee2018snip} leverages connection sensitivity to retain critical connections and remove less significant ones, aiming to limit the loss due to pruning. On the other hand, \textit{GraSP}~\cite{wang2020grasp} focuses on preserving the gradient flow through the network by scoring weights based on the Hessian-gradient product. \textit{SynFlow}~\cite{tanaka2020synflow}, being a data-agnostic pruning algorithm, emphasizes on preserving the total flow of synaptic strengths through the neural network to achieve Maximal Critical Compression during initialization.

\paragraph{Model Pruning in Federated Learning}
State-of-the-art research for model pruning in FL utilizes two main approaches: one involves the pruning mask being decided on the server side and the client only using this mask without changing it, and the other involves the collaborative selection of the pruning mask by both the user and the server side. \textit{PruneFL}~\cite{jiang2022pruneFL} follows the first approach, aiming to reduce the size and complexity of FL models without compromising accuracy. It achieves this by removing less important parameters and connections in each layer of the neural network based on their magnitude and importance scores. But the pruning mask is only decided by the server without considering user-side local representations.
\textit{LotteryFL}~\cite{li2020lotteryfl} belongs to the second approach, introducing a method that allows users to maintain local representations by selecting a subset of the global network using personalized masks. However, the system of sparse models it produces performs well only on local datasets.
\textit{FedDST}~\cite{bibikar2022feddst} involves dynamic sparse training on both the client and server sides, extracting and training sparse sub-networks. These highly sparse matching sub-networks are then transmitted to reduce both local training and communication costs in FL.
Despite the importance of privacy in FL, none of these methods have quantified the amount of privacy leakage after using model pruning. Hence, it is crucial to conduct theoretical and empirical analyses to measure their privacy implications.

\paragraph{Reconstruction Attacks in FL}
In terms of privacy leakage, communicating gradients throughout the training process in federated learning can reveal sensitive information about participants. \textit{Deep leakage from gradients} (DLG) \cite{zhu2019deep} demonstrated that sharing the gradients can leak private training data, including images and text. \textit{Improved DLG} (iDLG) \cite{zhao2020idlg}, presented an analytical procedure to extract the ground-truth labels from the shared gradients and showed the advantages of iDLG over DLG. \textit{Inverting Gradients} \cite{geiping2020inverting} introduced cosine similarity as a cost function in their reconstruction attacks and employed total variation loss $\mathcal{L_{TV}}$ as an image prior. 
\textit{Inverting Gradients} \cite{geiping2020inverting} and \textit{GradInversion} \cite{yin2021see} demonstrated the capability to reconstruct high-resolution images with increased batch sizes. The reconstruction quality of the attack was improved in \textit{Reconstruction From Obfuscated Gradient} (ROG) \cite{yue2022gradient}. ROG \cite{yue2022gradient} proposed conducting the reconstruction optimization in low dimensional space and trained a neural network as a postprocessing module.

While these privacy attacks have demonstrated the privacy vulnerabilities of standard federated learning, none of them has considered pruned federated learning. Therefore, we do not currently know how effective existing attacks are against the sparser models from pruned FL.

 \begin{figure}[t]
      \centering
      \includegraphics[scale=0.44]{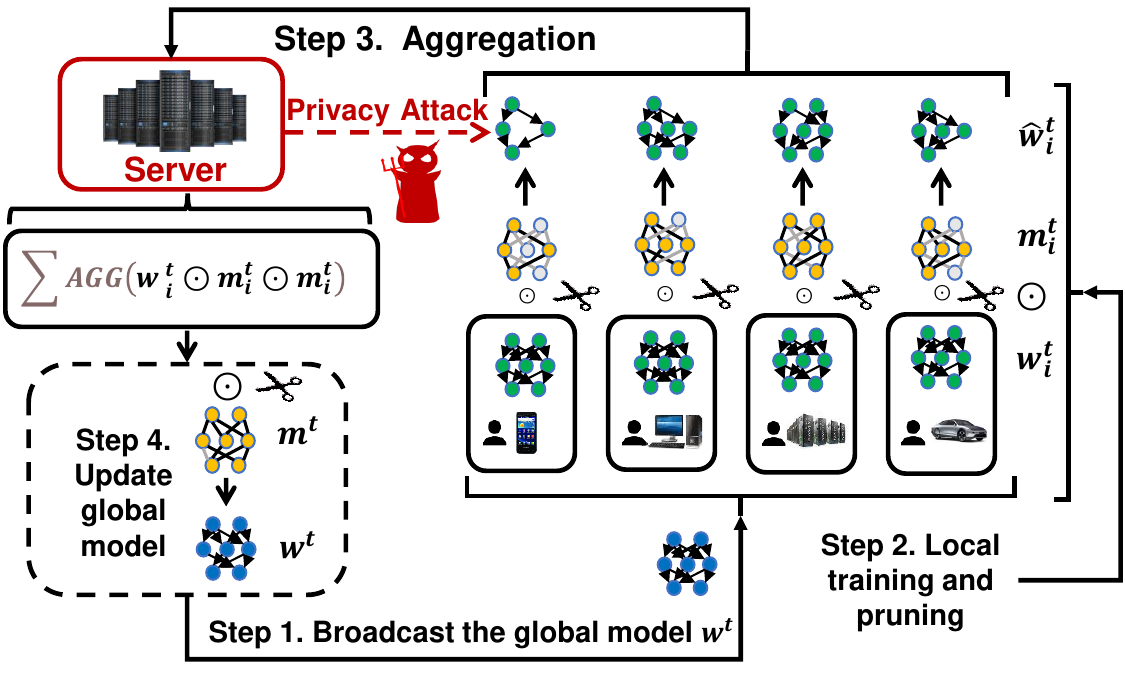}
      \caption{Setup: FL  with model pruning. Privacy attacks (by the honest-but-curious server)  and defense (at the clients).}
      \label{fig:overview}
     \end{figure}

\section{Problem Formulation}
 We consider a general system setup: FL with multiple users and one server, employing any model pruning scheme on the clients and/or the server, where privacy attacks and defense mechanisms are incorporated. This is depicted in Figure~\ref{fig:overview} and formalized in Algorithm~\ref{alg:pruning_with_attacks}.

The goal is to train a global model $F(\mathcal{D}, \bm{w})$ through FL. 
At the beginning of round $t$, user $i$ has local model parameters $w_i^t$; the server broadcasts the most recent global model to all participating users. The round proceeds as follows:

\noindent \textbf{Local Training and Pruning}: Each user $i$ with the local dataset ($\mathcal{D}_{i}$,$y_i$) utilizes the pruning strategy $\mathds{P}_{i}(\cdot)$ with the base pruning rate $p$ on its trained local model $F(\mathcal{D}_{i}; \bm{w}_{i}^{t})$ :
\begin{equation}
\label{eq:prune_training}  
    \hat{\bm{w}}_{i}^{t} := 
    \left ( \bm{w}^{t} - r\frac{\partial \ell_{i}\left(F({\mathcal{D}_{i}, \bm{w}^{t} \odot m^{t}}), y_i\right)}{\partial \bm{w}^{t}} \right ) \odot m^{t}_{i} 
\end{equation} 
Where $m_{i}^{t} = \mathds{P}_{i}(\bm{w}_{i}^{t})$ and $\odot$ is the Hadamard product. The base pruning masks $m_{i}^{t}$ generated by $\mathds{P}(\bm{w}^{t})$ can vary for different users and are permitted to be time-varying, allowing for dynamic adjustments of pruned global and local models throughout the  FL process.
Then user $i$ transmits its pruned local model parameters $\hat{\bm{w}}_{i}^{t}$ 
back to the server.  We refer to pruning at this stage as {\em base} (or {\em original}) pruning with rate $p$; it can be any state-of-the-art pruning FL scheme.

\noindent \textbf{Aggregation}: The server receives the model updates from all participating users and aggregates them to update the global model. It is important to note that we assume the base FL setting, without the implementation of secure aggregation~\cite{so2022lightsecagg} during this aggregation process.

\noindent \textbf{Privacy Attacks}: We assume an {\em honest-but-curious} server, who receives and stores model updates from each user and whose goal is to reconstruct users' training data. The server has full visibility of all masked local models and launches privacy attacks upon receiving an update from a target user.%
 We consider reconstruction attacks that invert gradients, including Deep-Leakage-from-Gradients (DLG) ~\cite{zhu2019deep}, Gradient Inversion (GI) ~\cite{geiping2020inverting}, and a custom attack SGI described later.

\begin{algorithm}[t] 
	\small
    \DontPrintSemicolon
    \SetKwFunction{UserUpdate}{UserUpdate}
    \SetKwFunction{PrivacyAttack}{PrivacyAttack}
    \SetAlgoNoEnd
	\SetAlgoNoLine
	Given: $T$ number of global training rounds; Users indexed by $i$; $E$ number of local epochs;
  $r$ is the learning rate; the server aims to reconstruct the local data of target user $I$.\\
	{\bf Server executes:}\\
    \SP Initialize $\bm{w}^0$\\
	\SP \For{each round t = 1, ... $T$} {
		\SP 	\For{each user $i$ in parallel} {
			\SP 		$\hat{\bm{w}}_{i}^{t} \leftarrow$ \UserUpdate$( \bm{w}^{t-1}, {m}^{t-1}, i)$\\
			\SP \If{$i \in $ {\it target $I$}}{
			\SP $\nabla{\hat{\bm{w}}_{i}^{t}} \leftarrow \hat{\bm{w}}_{i}^{t} - \hat{\bm{w}}_{i}^{t-1}$ \\
			\SP \PrivacyAttack($F(\mathcal{D}_{i}; \hat{\bm{w}}_{i}^{t})$, $\hat{\bm{w}}_{i}^{t}$,
   $\nabla{\hat{\bm{w}}_{i}^{t}}$)} %
			} %
        \SP ${\bm{w}}^{t} \leftarrow \left(\sum_{i=1}^I \frac{\parallel\mathcal{D}_{i}\parallel}{\parallel\mathcal{D}\parallel}
        \hat{\bm{w}}_{i}^{t}\right)\odot {m}^{t}$;\\
       }
	\BlankLine
	{\bf function \UserUpdate($\bm{w}^{t-1}, {m}^{t-1},i$):}\\
 	\SP  \For{each local epoch e = 1...$E$}{%
       \SP $\hat{\bm{w}}_i^{t} \leftarrow$ $\hat{m}_{i}^{t} \odot$ Eq.~(\ref{eq:prune_training}) 
        \label{line:defense}
	}

	\SP  \Return{${\hat{\bm{w}}}_i^{t}$ to server}
\caption{Pruning in FL under Privacy Attacks}
\label{alg:pruning_with_attacks}
\end{algorithm}

Intuitively, model pruning results in fewer parameters thus exposing less information to the server. However,  the server can still launch an attack to infer private data $\mathcal{D}_{i}$. In order to assess the level of privacy protection provided by model pruning methods in FL, we set out to answer the following questions: %
(1) how much information does the pruned FL model expose about the local dataset of a user? and (2) how can we further enhance privacy protection in pruned FL?

\noindent \textbf{Potential Defense:} W.r.t. the second question, we are interested in defense strategies that further enhance privacy in FL settings, specifically {\em via local pruning}, implemented in Algorithm~\ref{alg:pruning_with_attacks} Line~\ref{line:defense}. This family of defense strategies is universal in the following sense: defense mask $\hat{m}_i$ can always be applied locally and personalized for user $i$, after {\em any} base local pruning, $m_i$, and before sending the updates to the server, which can apply {\em any} privacy attack.
We refer to pruning at this stage as {\em defense} pruning with rate $\hat{p}$, to distinguish it from  {\em base} pruning above, and we propose \algoname for it.

It is worth noting that we consider existing privacy defenses known in the literature, including such as secure aggregation (SA) and differential privacy (DP)~\cite{mcmahan2018dp}, out-of-the-scope of this paper; these ideas are orthogonal and can be complementary/combined with our pruning-based defense in the future.

\section{Theoretical Analysis}
\label{sec:theoretical}
In this section, we provide a theoretical quantification of privacy leakage inherent in model pruning within the context of FL. We aim to  explore the server's ability to infer information about an individual user's local dataset $\mathcal{D}_{i}$.
We employ mutual information as a foundational metric for quantifying privacy leakage. This concept, rooted in information theory and Shannon entropy, captures the interdependence between two random variables.
The use of mutual information is ideal for our privacy analysis, as it enables the measurement of privacy leakage considering non-linear relationships between variables, and thus offers a measure of true dependence.
Normalized mutual information is applicable across various domains and tasks~\cite{belghazi2018mutual,bachman2019learning}, as well as across all pruning schemes and attack scenarios. It was recently used to also quantify the privacy due to FL aggregation ~\cite{elkordy2023much}.

In this work, we seek to quantify how much information the server can infer about the private data $\mathcal{D}_{i}$ of user $i$ over $T$ global training rounds, based on the pruned updates $\left \{ \bm{w}^{t}_{i} \odot m^{t}_{i}  \right \}_{t \in \left [ T \right ] }$ submitted by user $i$:
\begin{align}
\label{eq:multi_round}
    \mathbf{I}_{i} = \mathbf{I} \left ( \mathcal{D}_{i} ; \left \{ \bm{w}^{t}_{i} \odot m^{t}_{i}  \right \}_{t \in \left [ T \right ] }\right ) 
\end{align}
where $\mathbf{I}$ represents the mutual information, $\bm{w}_{i}^{t} \in \mathds{R}^{d}$ is the local model weights of user $i$ at round $t$.%
 To that end, we first characterize the privacy leakage for a {\em single round $t$}:
\begin{align}
    \mathbf{I} \left ( \mathrm{x}_{i}^{t} ;  \bm{w}_{i}^{t} \odot m_{i}^{t} \middle| \left \{ \bm{w}^{k}_{i} \odot m^{k}_{i}  \right \}_{k \in \left [ t-1 \right ] } \right)
\end{align}
where
$\mathrm{x}_{i}^{t} = \frac{\partial \ell_{i}({\hat{\bm{w}}_{i}^{t-1};\mathcal{D}_{i}})}{\partial \bm{w}}=  \frac{1}{B}\sum_{b \in \mathbf{B}_{i}^{t}}g_{i}\left (\bm{w}_{i}^{t},b\right )$
, $g_{i}(\bm{w}_{i}^{t})$ denotes the stochastic estimate of the gradient of the local loss function for user $i$, computed based on a random sample $b$, and $B$ is the batch size. Also, we define $\hat{g_{i}}\left (\bm{w}_{i}^{t},b\right) \in \mathds{R}^{d^{*}}$ is the largest sub-vector of the $g_{i}\left (\bm{w}_{i}^{t},b\right)$ with non-singular covariance matrices. %
\begin{assumption}
\label{as:guassian}
The random vector $\hat{g_{i}}\left (\bm{w}_{i}^{t},b\right )$ has non-singular covariance matrix $\Sigma_{i}$ and mean 0.
\end{assumption}
\begin{assumption}
\label{as:pruning_ratio}
$p_{i}$ is the base pruning rate of user $i$, remaining unchanged throughout the training process.
\end{assumption}
Under these two mild assumptions, we obtain our main theoretical result; the proof is deferred to the Appendix.
\begin{theorem}[Single Round Leakage] Under Assumption~\ref{as:guassian} and ~\ref{as:pruning_ratio}, we have an upper bound of $\mathbf{I}_{i}^{t}$ for a single round:
\begin{align}
     & \mathbf{I}_{i}^{t} \leq 1 - \frac{p_{i}-1}{2\ln2} + 2\log\frac{1}{B}
     +  2\Delta
    + d^{*}\log \left (2 \pi e\right)
\end{align}
where $\Delta = \log\left|\det(\Sigma_{i}^{-\frac{1}{2}})\right|$.
\label{th:single_round}
\end{theorem}

We see that 
    The upper bounds on information leakage decrease with larger pruning rate ($p$) and batch sizes ($B$), while they increase with the increase in $d^{*}$. Notably, the impact of model size ($d$) on information leakage is not linear. %

\begin{corollary}[Multiple Rounds Leakage] Continuing with Theorem~\ref{th:single_round}, we can upper bound $\mathbf{I}_{i}$ after $T$ global training rounds as follows:
\begin{align}
      \mathbf{I}_{i} \leq T\left( 1 - \frac{p_{i}-1}{2\ln2} + 2\log\frac{1}{B}
     +  2 \Delta
    + d^{*}\log \left (2 \pi e\right)\right)
\end{align}
\label{co:multi_round}
\end{corollary}
\vspace{-10pt}
where $\Delta = \log\left|\det(\Sigma_{i}^{-\frac{1}{2}})\right|$. We see that increasing the number of global training rounds ($T$) results in a proportional rise in information leakage in the user's local training model. %

The upper bounds are powerful in that they apply to {\em any} pruning scheme, and {\em any} privacy attack. In practice, the bounds may or may not be tight compared to the actual information leakage, depending on the model, pruning method,  privacy attack, and dataset.
We have compared the analysis (in terms of normalized mutual information) to practical evaluation (based on PSNR) and we found that the metrics agree; some results are presented in the next sections and some are deferred to the appendix.

\section{Privacy Attacks against Model Pruning in FL}
In this section, we conducting a comprehensive evaluation to quantify the privacy-preserving properties of different existing pruning methods in FL. By evaluating these methods under real-world conditions, we seek to evaluate their effectiveness across different datasets and under true attack scenarios, and to refine and validate our  theoretical analysis.

\subsection{Experimental Setup}
\subsubsection{Datasets and Models}
We evaluate models Conv-2~\cite{caldas2019leaf} and VGG-11~\cite{simonyan2014very} ) along with their corresponding datasets FEMNIST~\cite{caldas2019leaf} and CIFAR-10~\cite{krizhevsky2009learning}), which are commonly employed in FL studies.

\subsubsection{Implementation details}
In each round of training, we randomly select 10 clients from a pool of 193 users for the FEMNIST dataset and 100 clients for the CIFAR-10 dataset. The training process involves 5000 iterations, with a batch size of 20 for the FEMNIST dataset and a batch size of 1 for the CIFAR-10 dataset with local updates and a learning rate of 0.25. More implementation details are in the Appendix.

\paragraph{Privacy Attack Algorithm}
We first implement the classic Gradient Inversion(GI) attack~\cite{geiping2020inverting} from the DLG attack family. However, due to the unique sparsity patterns introduced by model pruning FL, we identify the potential for further optimizing this attack to specifically exploit the sparsity. Consequently,
we introduce an advanced attack tailored for the context of model pruning in FL, the Sparse Gradient Inversion (SGI) attack. This attack is designed with the goal of recovering the pruning mask within the pruned model. Moreover, its adaptability allows seamless integration with existing privacy attacks in FL, thereby amplifying their effectiveness.
The algorithm for the SGI attack is provided in Algorithm~\ref{alg:SGI}. In Line~\ref{alg:SGI_mask}, based on user $i$'s local model weights $\bm{w}_{i}^{t}$, the server attempts to first recover the pruning mask of user $i$, $m_i$. The recovered mask is then integrated into the calculation of the gradient, allowing the server to obtain 
 $\nabla{\hat{\bm{w}}_{i}^t}$ that closely approximates the true gradients of user $i$.

The comparison of attack algorithms, deferred to the Appendix, shows that the SGI outperforms the GI attack across all scenarios.%
In the rest of the paper, we use the SGI attack as baseline, with learning rate 0.01 and 10,000 iterations. %

\begin{algorithm}[t]
	\small
	\SetAlgoNoEnd
	\SetAlgoNoLine
	\DontPrintSemicolon

	{\bf Input:} $F(\mathcal{D}_{i}; \hat{\bm{w}}_{i}^{t})$: model at round $t$ from targeted user $i$; learning rate $\eta$ for inverting gradient optimizer; $S$: max iterations for attack; $\tau$: regularization term for cosine loss in inverting gradient attack; $m_{rec}$:the recovery mask generated by sever; $d$:the model size\\
	{\bf Output:} reconstructed training data $(\mathcal{D}_{i},y_i)$ at round $t$\\
    \SP$m_{rec}[j]_{j\in d} = \begin{cases}
            1 & \text{ if } \hat{\bm{w}}_{i}^{t}[j] \neq 0 \\ 
           0 & \text{ if }  \hat{\bm{w}}_{i}^{t}[j] = 0
            \end{cases}$\\ \label{alg:SGI_mask} 
     \SP $\nabla{\hat{\bm{w}}_{i}^t} \leftarrow \left(\hat{\bm{w}}_{i}^{t} - \hat{\bm{w}}_{i}^{t-1}\right) \odot m_{rec}$ \\
    \SP Initialize $\mathcal{D}'_0 \leftarrow \mathcal{N}$(0,1), $y'_0 \leftarrow $ $ Randint(0, max(y))$\\
	\SP \For{s$\leftarrow$ 0,1, ... $S-1$} {
		\SP 	$\nabla{\bm{w}'_s} \leftarrow \partial{\ell({F(\mathcal{D}'_s, \hat{\bm{w}}_i^t), y'_s)}}$/$\partial{\hat{\bm{w}}_i^t}$\\
		\SP 	$\mathcal{L}'_s \leftarrow 1 - \frac{\nabla{\hat{\bm{w}}_{i}^t}\cdot\nabla{\bm{w}'_s} }{\parallel\nabla{\hat{\bm{w}}_{i}^t}\parallel\parallel\nabla{\bm{w}'_s} \parallel} + \tau$   \tcp*{cosine  loss}
	    \SP 	$\mathcal{D}'_{s+1} \leftarrow \mathcal{D}'_{s} - \eta\nabla_{\mathcal{D}'_{s}}\mathcal{L}'_s$, $y'_{s+1} \leftarrow y'_s - \eta\nabla_{y'_s}\mathcal{L}'_s$
      }
	\SP \Return{$\mathcal{D}'_{S}, y'_{S}$}
	\caption{Sparse Gradient Inversion (SGI) Attack}
\label{alg:SGI}
\end{algorithm}

\subsubsection{Evaluated Pruning Methods} %
We subject a set of well-established pruning methods to the privacy attack to assess their vulnerabilities and effectiveness in protecting user privacy. The pruning methods under evaluation include \textit {Random pruning}, Snip~\cite{lee2018snip}, SynFlow~\cite{tanaka2020synflow}, GraSP~\cite{wang2020grasp}, FedDST~\cite{bibikar2022feddst}, and PruneFL~\cite{jiang2022pruneFL}, all of which were introduced in Related Work.

\begin{figure*}[!htbp]
  \centering
  \begin{subfigure}[b]{0.23\textwidth}
    \centering
    \includegraphics[width=\linewidth]{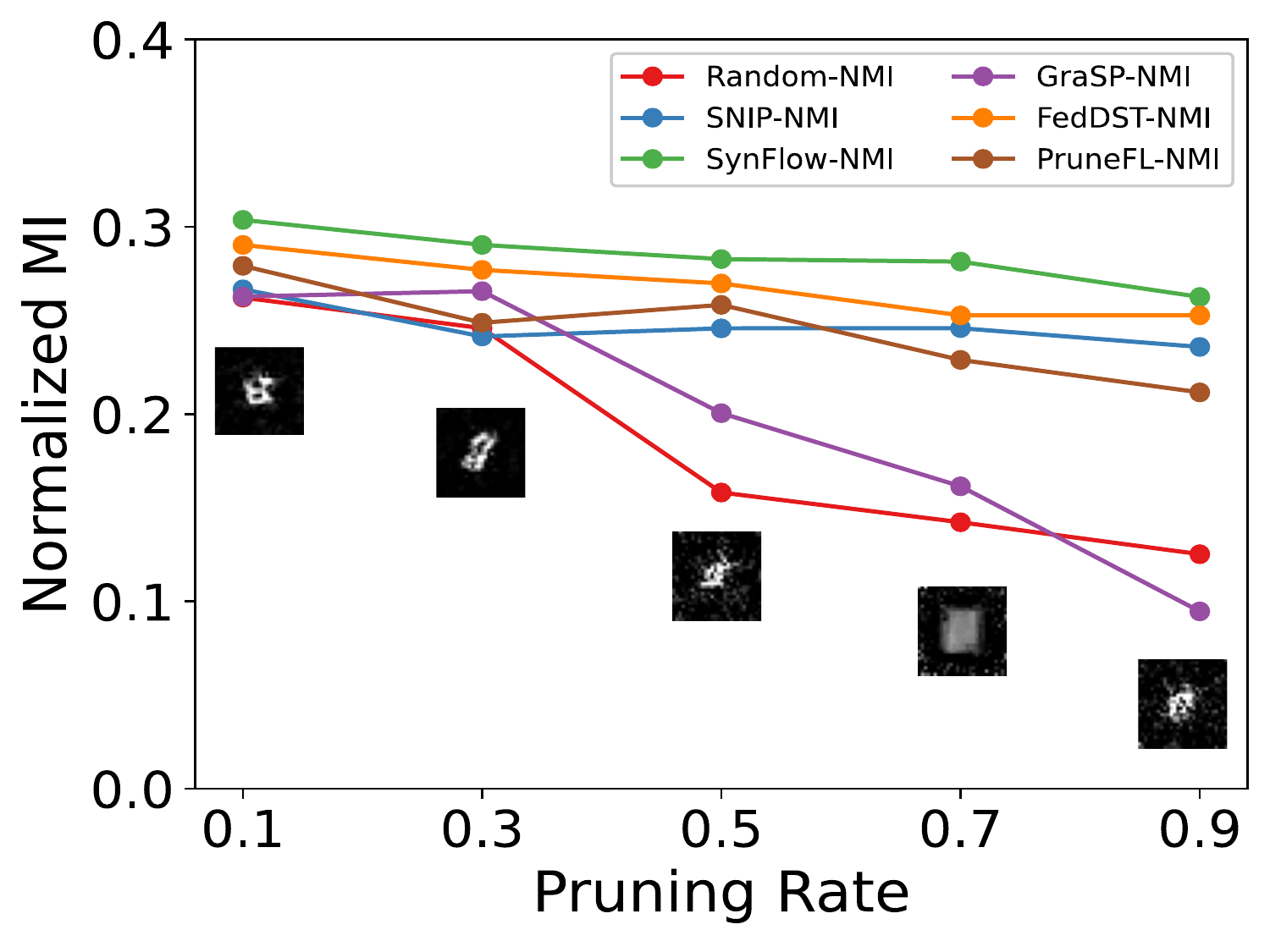}
    \caption{Varying the pruning rate $p$.}\label{fig:pruning_rate_femnist}
  \end{subfigure}
  \hfill
  \begin{subfigure}[b]{0.23\textwidth}
    \centering
    \includegraphics[width=\linewidth]{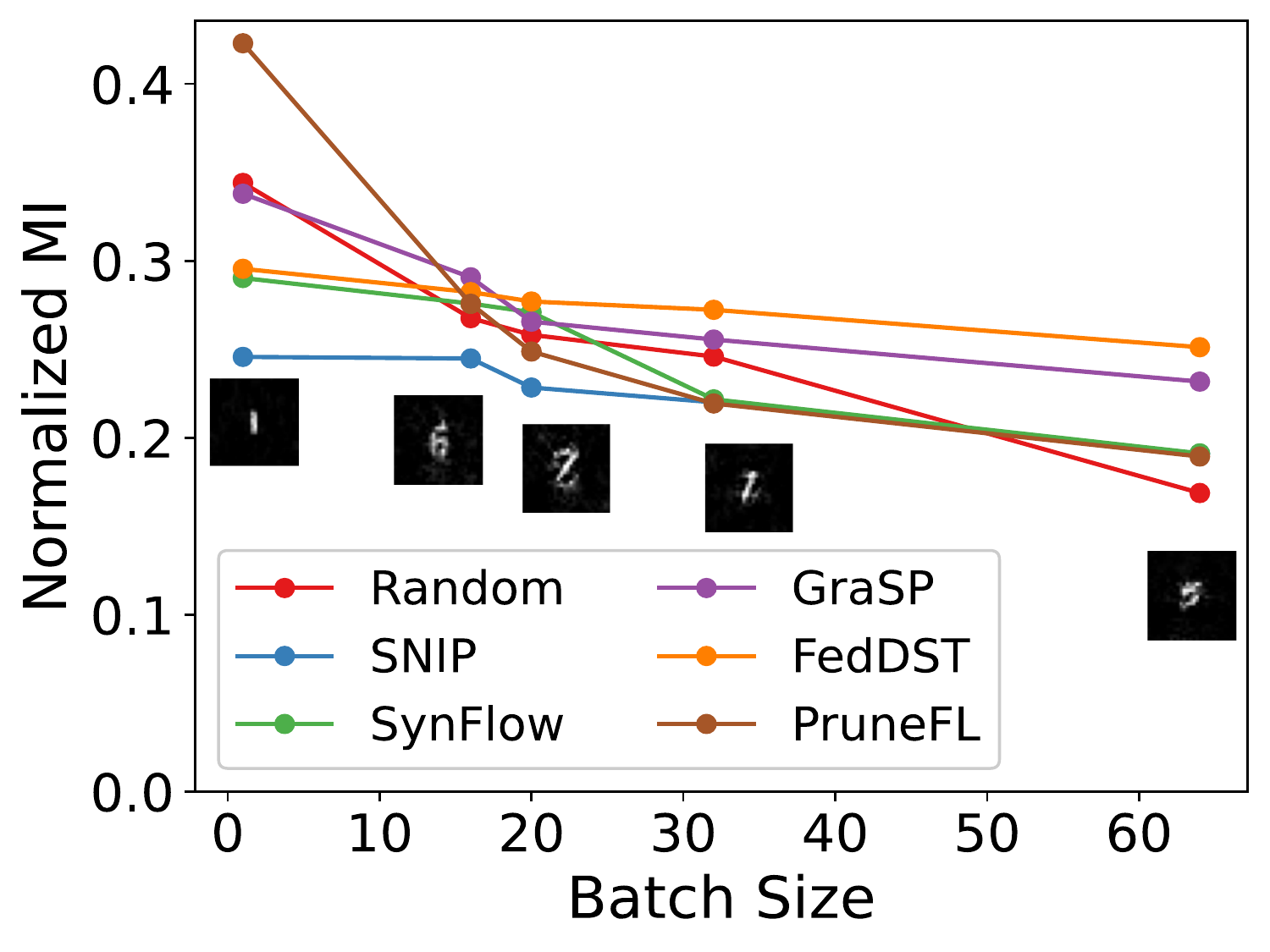}
    \caption{Varying the batch size $B$.
    }
\label{fig:batch_size_femenist}
  \end{subfigure}
  \hfill
  \begin{subfigure}[b]{0.23\textwidth}
    \centering
    \includegraphics[width=\linewidth]{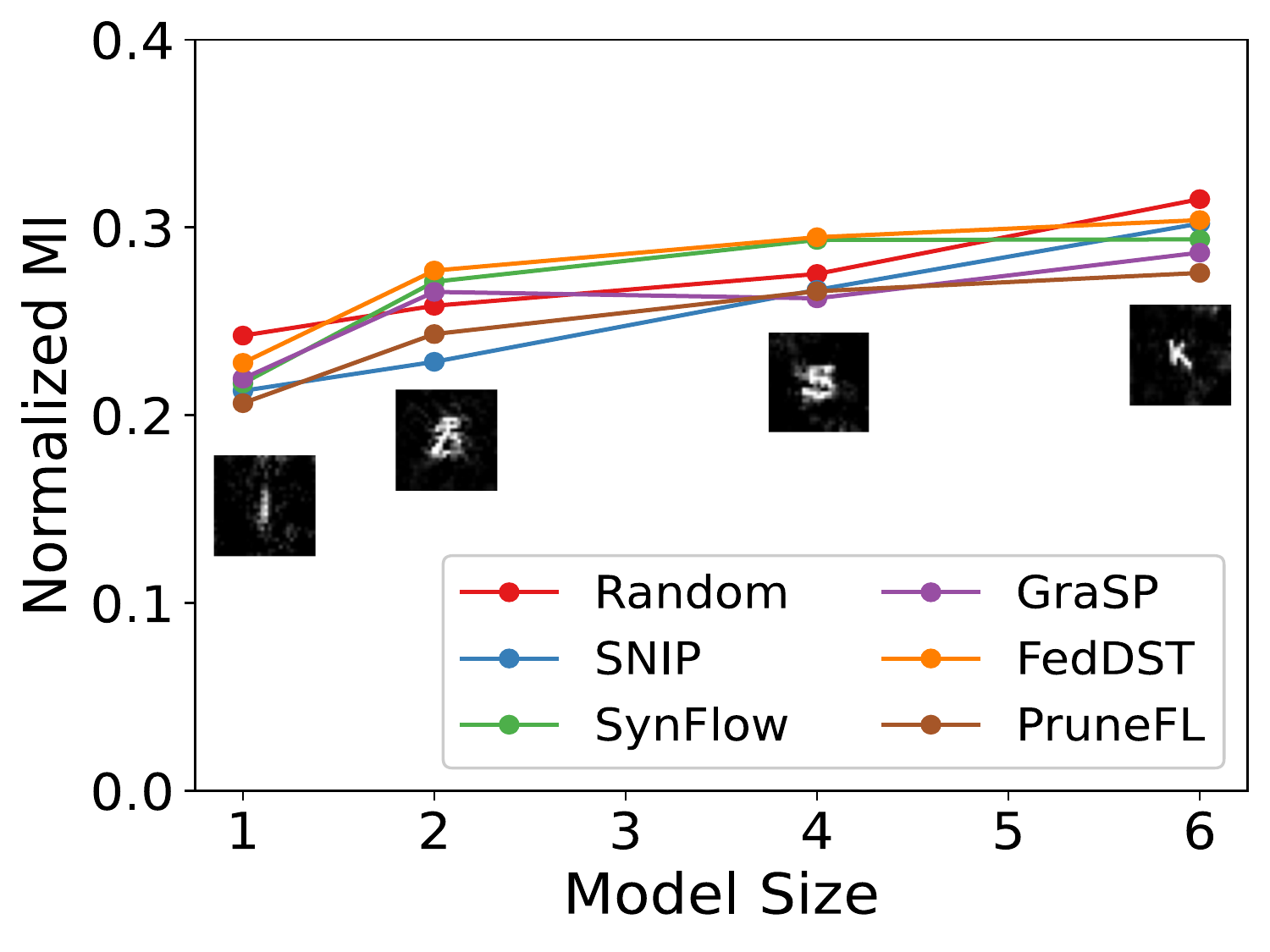}
    \caption{Varying the model size $d$.
    }
\label{fig:model_size_femnist}
  \end{subfigure}
  \hfill
  \begin{subfigure}[b]{0.23\textwidth}
    \centering
    \includegraphics[width=\linewidth]{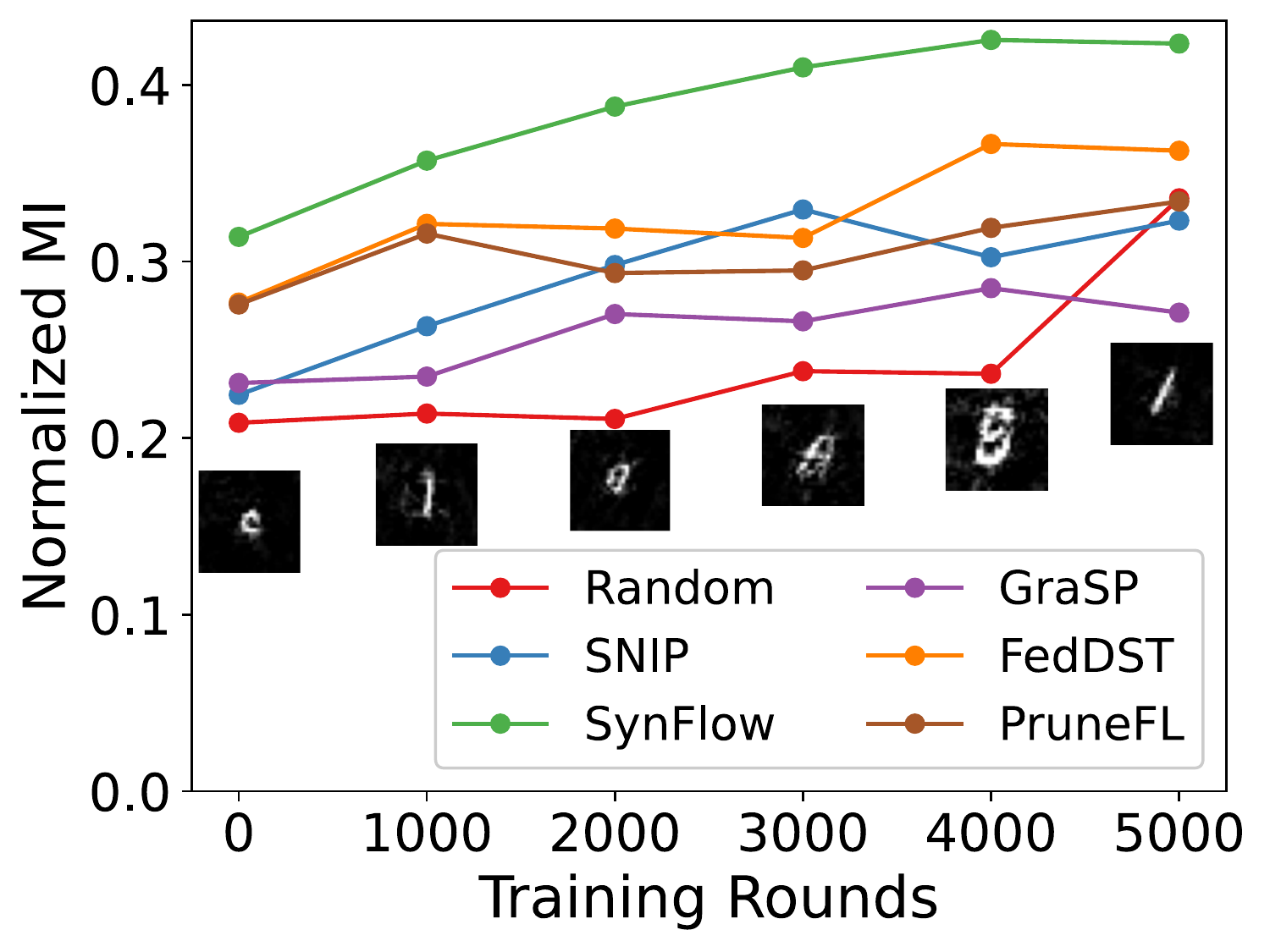}
    \caption{Varying the training round $T$ }
\label{fig:training_round_femnist}
  \end{subfigure}

 \caption{Impact of varying pruning rate, batch size, model size, and training rounds on privacy leakage using Normalized Mutual Information (NMI) under a Sparse Gradient Inversion (SGI) attack for six pruning methods: Random, SNIP, SynFlow, GraSP, FedDST, and PruneFL in FEMNIST dataset}\label{fig:attack_all_methods}
\end{figure*}

\subsubsection{Performance Metrics}
To assess the effectiveness of the evaluated pruning methods, we employ the following  metrics:

\noindent\textbf{Privacy Metrics:} We use the Normalized Mutual Information (NMI) \footnote{For two random vectors $U$, $V$ $\in \mathcal{R}^N$, 
their MI is $ I(U;V) = \sum_{i=1}^{|U|} \sum_{j=1}^{|V|}\frac{|U_{i}\cap V_{j}|}{N} \log\left(\frac{N|U_{i}\cap V_{j}|}{|U_{i}||V_{j}|}\right)$. NMI is defined as $\frac{I(U;V)}{\mathrm{mean}(H(U), H(V))}$.} between the training data and the reconstructed data. Higher values of NMI indicate a higher risk of privacy leakage. %
We also use the practical and well-established PSNR metric for images. A comparison of the two and supplemental experiments can be found in the Appendix. 
    
\noindent\textbf{Utility Metric:} We use model accuracy (ACC) to assess model performance. Ideally, we want improved privacy, without significantly compromising accuracy.

\subsection{Attack Performance}
We conduct empirical evaluations to assess how FL parameters influence privacy leakage in model pruning under privacy attacks. The results are in line with our theoretical findings. %

\subsubsection{Impact of Pruning Rate ($p$)}

Figure~\ref{fig:pruning_rate_femnist} illustrates the impact of varying pruning rates $p$ from 0.1 to 0.9 on privacy leakage using NMI on FEMNIST dataset. Consistent with our Theorem~\ref{th:single_round}, we observe that higher pruning rates lead to a proportionally decreasing trend in NMI. This reduction indicates a lowered risk of privacy leakage due to the fewer parameters present in the pruned model.  
 
\subsubsection{Impact of Batch Size ($B$)}
Figure~\ref{fig:batch_size_femenist} shows the impact of varying batch size $B$ on privacy leakage using NMI on the FEMNIST dataset, where $B$ is selected from the set {1, 16, 20, 32, 64}.
It demonstrates that larger batch sizes increase the privacy protection of the FL process as they enable more data points to be added during local training, which can help in obscuring the data characteristics of a particular user.   

\subsubsection{Impact of Model Size ($d$)}
We consider 4 architecture  of the models:
Conv-1, Conv-2, Conv-4, and Conv-6, featuring 1, 2, 4, and 6 convolutional layers, along with 2 linear layers. These models encompass 3,815,566, 6,603,710, 13,657,406, and 25,324,350 parameters, respectively. 
Figure~\ref{fig:model_size_femnist} depicts the relationship between the model size ($d$) and the corresponding increase in information leakage. The x-axis in the figure represents the number of convolutional layers in each model. This observation can be attributed to the fact that larger models possess a higher number of parameters, thereby amplifying gradient information leakage.

\subsubsection{Impact of Communication Round ($T$)}
In Figure~\ref{fig:training_round_femnist}, the depiction of training rounds $T$ on information leakage reveals a non-linear increase. This observation aligns with Corollary~\ref{co:multi_round}, which provides an upper bound not strictly indicative of actual performance. The algorithms surpass this bound, allowing for potential non-linear growth with increasing rounds, without contradicting the established theorems.

Due to the page limits, we only present results on the FEMNIST dataset, the results for other datasets are provided in the Appendix. Notably, the observed impacts of the parameters $p$,$B$,$d$, and $T$ remain consistent across all of these datasets. These evaluations demonstrate that pruning alone lacks sufficient defensive capabilities against privacy attacks, as even under a high pruning rate ($p=0.9$), the server is able to reconstruct local data from the pruned model.

\section{Pruning as a Privacy Defense for FL}
\begin{figure}[t!]
    \centering  \includegraphics[scale=0.5]{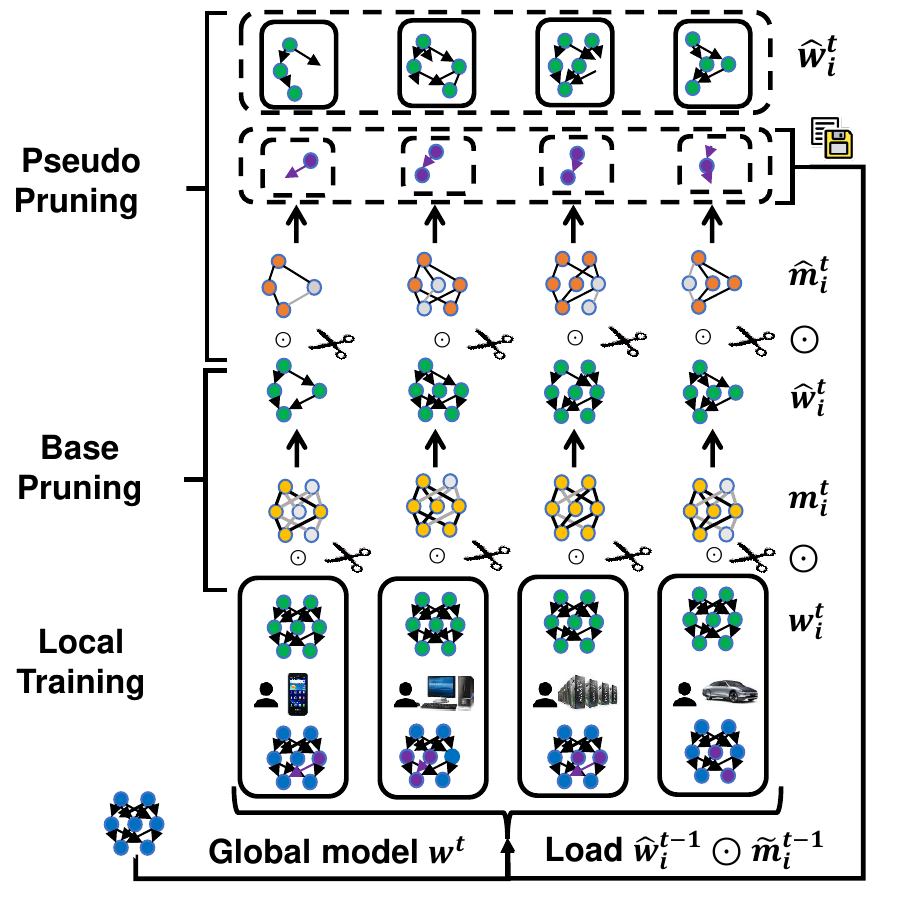}
    \caption{Illustration of the \PseudoDefense Process}
    \label{fig:pseudo}
\end{figure}

Pruning has been typically designed with model accuracy and model size in mind. Here, we reflect on evaluation results and obtain valuable insights into the choices and parameters of pruning in FL that affect the privacy provided by pruning. Then, we use these insights to design a pruning mechanism, specifically with privacy in mind, that is applied locally, {\em after} the base pruning FL scheme.

\subsection{Insights from Evaluation}

\noindent \textbf{Insight 1: Largest Weights Matter.}
Pruning weights with large gradients improves privacy the most, but also hurts model accuracy the most.

Pruning alone is inadequate to defend against privacy attacks. Indeed, pruning criteria as used in \textit{PruneFL} and \textit{FedDST},  retain the weights with the largest gradients to preserve valuable information for the model, which can then be exploited by DLG-type of attacks.

This insight implies that the defense should strategically prune certain weights with large gradients, in addition to those pruned by the base pruning scheme. 
We explored three options for weights to prune:
(1)  weights with the top-$k$ largest gradients (denoted as \textit{Largest}), (2) random weight pruning (denoted as \textit{Random}), and (3) a hybrid approach combining the \textit{Largest} and \textit{Random} (denoted as \textit{Mix}). We compared the three strategies for the same pruning rate $p$ set (0.3) on top of the base method (\textit{PruneFL}) and more details are in the Appendix. The results in Figure \ref{fig:prelim_defense} within the left dashed circle show a tradeoff between privacy 
and accuracy. 
It illustrates that pruning more weights with large gradients improves privacy due to valuable information reduction, 
but also significantly reduces accuracy, as observed in the 'Largest' method. Therefore, to preserve privacy while simultaneously improving accuracy, we need an additional mechanism to maintain model accuracy.

\noindent \textbf{Insight 2: \PseudoDefense.} 
Users can prune weights with
large gradients when they communicate with the server (which helps privacy) {\em and} still keep these weights for their local training (thus helps accuracy).

We refer to this idea as \PseudoDefense, and its process is illustrated in Figure~\ref{fig:pseudo}.%
Users locally save the weights designated for pruning via the defense mask, but do not transmit them to the server.
Specifically, before sending the weights $\hat{\bm{w}}_{i}^{t}$ pruned by the base prune strategy
to the server, %
user $i$ conducts \PseudoDefense with defense mask $\hat{m}_{i}^{t}$.  The weights marked for pruning $\hat{\bm{w}}_{i}^{t} \odot {\tilde{m}}^{t}_{i}$ are withheld locally; only the weights after the defense pruning $\hat{\bm{w}}_{i}^{t} \odot {\hat{m}}^{t}_{i}$, are transmitted to the server. Here, $\tilde{m}$ is the bit-wise complement matrix of $\hat{m}$. In the next round, after user $i$ receives the global model $\bm{w}^{t+1}$, the locally saved weights are loaded into the global model, serving as the local initial model. %

 \begin{figure}
   \centering \includegraphics[scale=0.35]{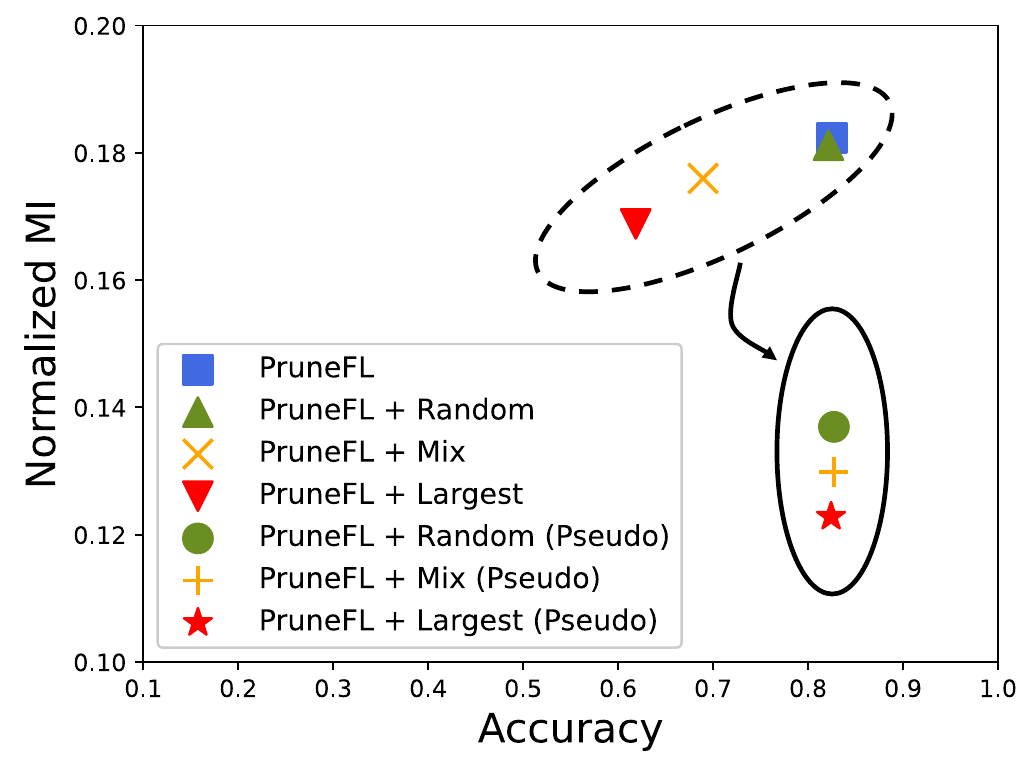}
     \caption{ Tradeoff analysis of privacy and accuracy with three defense strategies (\textit{Largest}, \textit{Random}, and \textit{Mix}), and their enhanced versions with \PseudoDefense (\textit{Pseudo}).
     }   \label{fig:prelim_defense}
 \end{figure}
Figure~\ref{fig:prelim_defense} shows that the model accuracy is notably better when using the same mask and \PseudoDefense than real pruning. 
Moreover, the evaluation shows that combining the two insights into %
\textit{Largest (Pseudo)} offers the most effective privacy guarantee among all the defenses discussed above. %
 Additional evaluation details are available in the Appendix. 
In the defense considered so far, the defense rate is manually selected and fixed. 

\noindent \textbf{Insight 3: Adapt.}
The defense pruning rate should adapt to the base pruning rate, so as to jointly optimize privacy and model accuracy.

The defense strategy should adapt to the changing dynamics of the training process, enabling a more effective equilibrium to be struck between the crucial trade-off of privacy and accuracy.
The defense pruning rate should be selected so that, taken together with the base pruning rate, it strikes a good trade-off between privacy and accuracy.%

\begin{figure*}[htbp!]
  \centering
  \begin{subfigure}{0.25\textwidth}
    \centering    \includegraphics[width=0.94\linewidth]{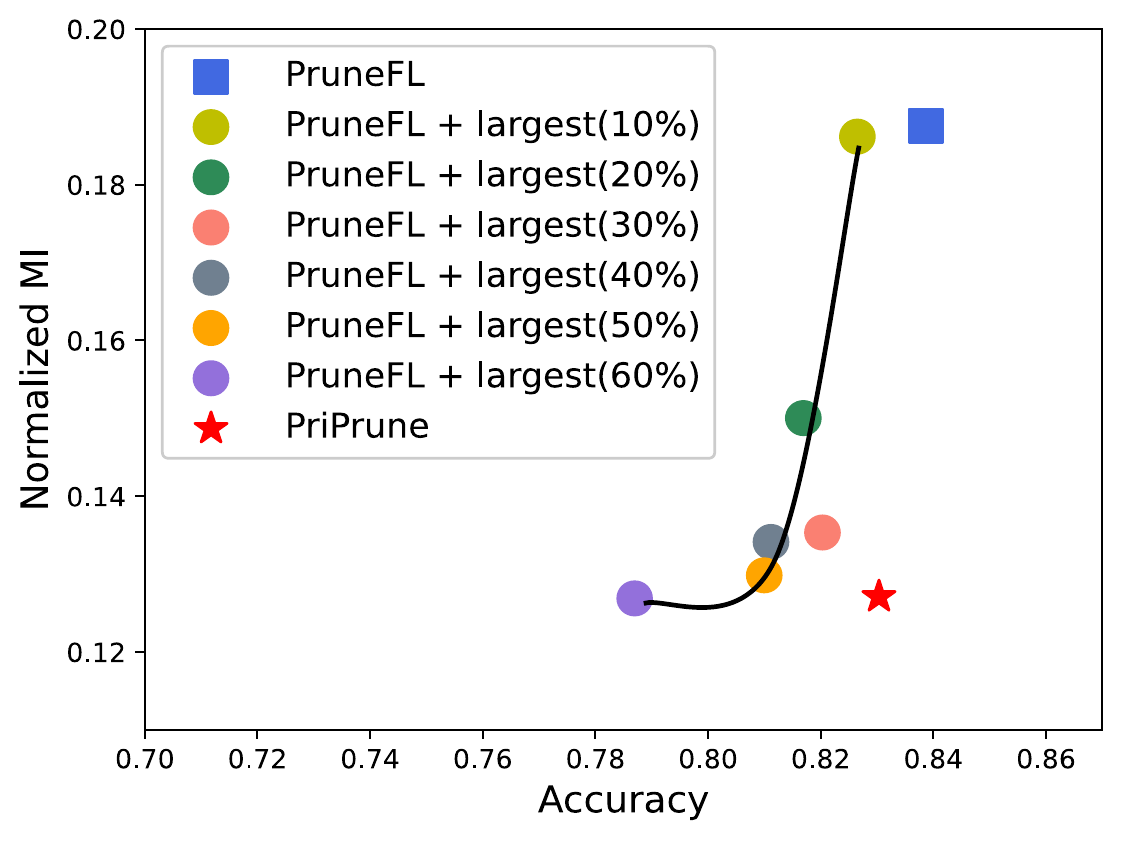}
    \caption{Privacy (NMI) vs. Utility (ACC) }    \label{fig:tradeoff_priprune_prunefl}
  \end{subfigure}
  \hfill
  \begin{subfigure}{0.23\textwidth}
    \centering
    \includegraphics[width=1.01\linewidth]{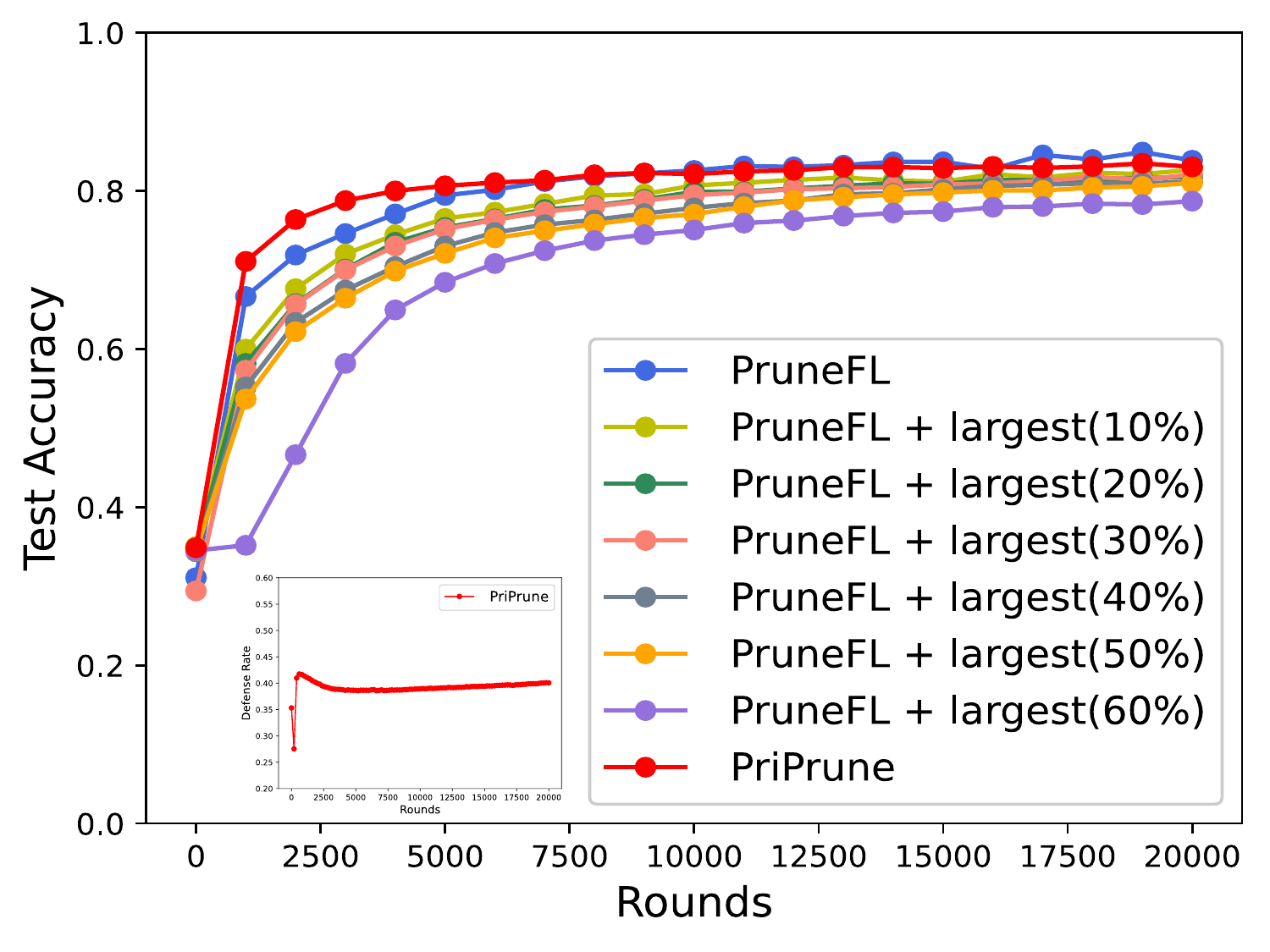}
    \caption{ACC with adaptive $\hat{p}$}
    \label{fig:priprune_prunefl_accuracy}
  \end{subfigure}
  \hfill
  \begin{subfigure}{0.23\textwidth}
    \centering
    \includegraphics[width=\linewidth]{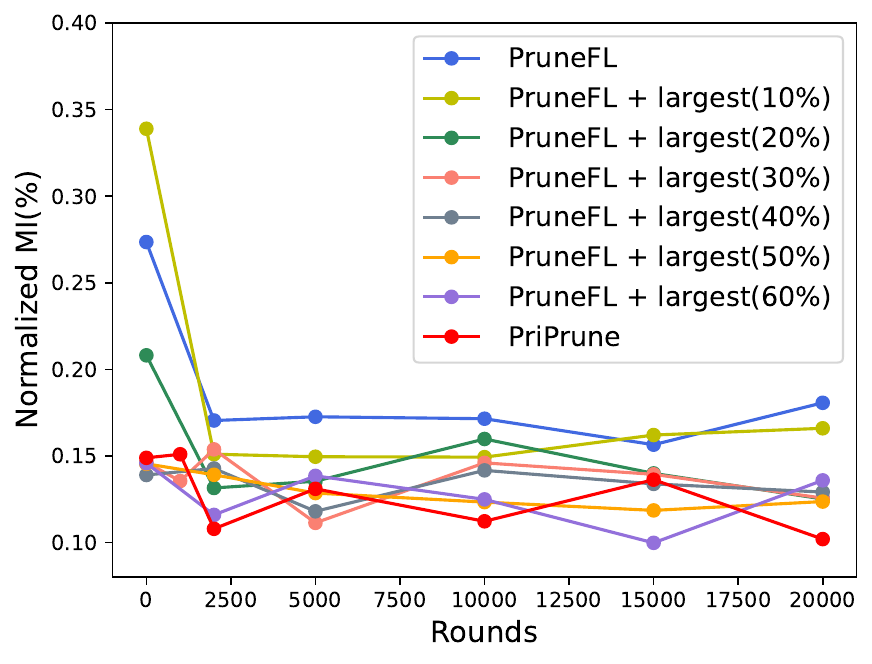}
    \caption{Privacy (NMI) over time}   \label{fig:priprune_prunefl_NMI}
  \end{subfigure}
  \hfill
  \begin{subfigure}{0.23\textwidth}
    \centering
    \includegraphics[width=\linewidth]{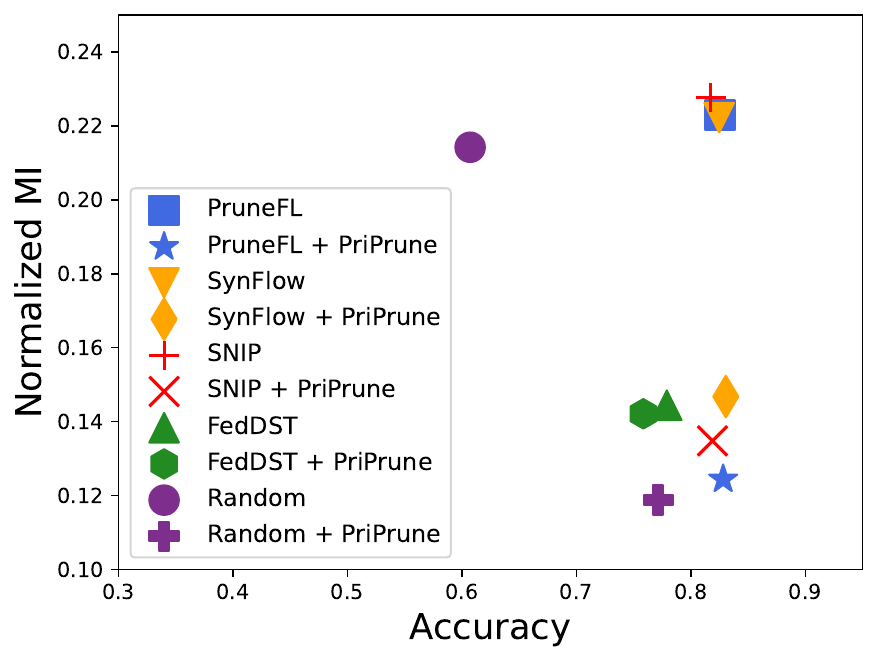}
    \caption{\algoname}    \label{fig:priprune_all_methods}
  \end{subfigure}
  \caption{Performance Evaluation of \algoname (as defense) with  PruneFL (as the base). (a) Privacy (in terms of NMI) vs. (Utility (ACC) trade-off for \algoname and ``Largest'' with different fixed defense rates. (b) Model Accuracy and Adaptive Defense rate $\hat{p}$ over time. (c) Privacy over time. (d) The whole trade-off analysis of integrating \algoname with various pruning mechanisms.
}.
 \label{fig:priprune_all}
\end{figure*}

\begin{algorithm}[t!]
\SetAlgoNoLine
\SetAlgoNoEnd
\caption{ \algoname }
\label{alg:adaptive_PriPrune}
\SetKwInOut{Input}{Input}
\SetKwInOut{Output}{Output}
\SetKwFunction{Pruning}{Pruning}
\SetKwFunction{Initialize}{Initialize}
\SetKwFunction{Save}{Save}
\SetKwFunction{Receive}{Receive}
\SetKwFunction{PruningStrategy}{PruningStrategy}
\SetKwFunction{UserUpdate}{UserUpdate}
\SetKwFunction{ServerAggregation}{Server Aggregation}

\BlankLine
{\bf \UserUpdate($i$, $\bm{w}^{t}$):}\\
    \SP \SP Load $\bm{w}_{i}^{t-1}$ from local storage of user $i$\\
    \SP \SP Initialize $\alpha_{i}^{t} = \alpha_{i}^{t-1}$\\ 
    \SP \SP \For{local epoch e $\in E$}
            {
            {\SP \SP Sample mask $\hat{m}_{i}^{t}$ by Eq.~\ref{eq:gumbel_soft} with $\alpha_{i}^{t}$ \\
            \SP \SP Composite batch parameter by:\\
            \SP \SP $\bm{w}_{i}^{t}[j]_{j \in d} = \begin{cases}
            \bm{w}_{i}^{t-1}[j] & \text{ if } \hat{m}[j]= 0 \\ 
            \bm{w}^{t}[j] & \text{ if } \hat{m}[j]= 1
            \end{cases}$, \\
            \SP \SP $\mathcal{L}_{local} \leftarrow$ Eq.(~\ref{eq:loss_func})\\
            \SP \SP  $\bm{w}_{i}^{t} \leftarrow \bm{w}_{i}^{t} - \eta\nabla\mathcal{L}_{local}(\bm{w}_{i}^{t},\alpha_{i}^{t})$ \\
            \SP \SP  $\alpha_{i}^{t} \leftarrow \alpha_{i}^{t} - \eta\nabla\mathcal{L}_{local}(\bm{w}_{i}^{t},\alpha_{i}^{t}) $ \\    
            }}
\SP \SP $\hat{m}_{i}^{t} \leftarrow argmax(\alpha_{i}^{t}, 1-\alpha_{i}^{t})$ \\
\SP \SP $\bm{\hat{w}}_{i}^{t} \leftarrow \bm{w}_{i}^{t} \odot \hat{m}_{i}^{t}$ \\
\SP \SP \Return{$\bm{\hat{w}}_{i}^{t}$} to server\\
\end{algorithm}

\subsection{The \algoname Defense Mechanism} %

We combine the aforementioned three insights, and we design \textbf{\algoname} -- an adaptive version of \PseudoDefense, which can dynamically adapt the defense mask $\hat{m}$ to optimize the accuracy and privacy trade-off throughout the FL training process. %
The objective of user $i$ is to minimize a loss function that captures that trade-off:
\begin{align}
\label{eq:loss_func}
    \min_{\hat{m}_{i}} \min_{w_{i}} \mathcal{L}_{local} = \lambda_{acc}\mathcal{L}_{acc}  + \lambda_{pri}\mathcal{L}_{pri} + \lambda_{sha}\sum\limits_{L, \hat{d_{l}}} \alpha
\end{align}

Here,  $\mathcal{L}_{acc}$ represents the local model training loss, %
$\mathcal{L}_{pri}$ represents the local privacy loss, and $\alpha$ represents the probability of not sharing parameters with the server, in \PseudoDefense. $\hat{d_{l}}$ represents the weights in layer $l \in L $ that are not shared. $\lambda_{acc}$, $\lambda_{pri}$ and $\lambda_{sha}$ are the weights for $\mathcal{L}_{acc}$, $\mathcal{L}_{pri}$ and the privacy regularization term, respectively.%

In \algoname, we optimize the defense mask $\hat{m}$ and model weights $w_{i}$ jointly through gradient descent based on our designed loss function. To overcome the discrete and non-differentiable nature of the defense mask $\hat{m}$, we employ Gumbel-Softmax sampling \cite{maddison2016concrete, jang2017categorical} to substitute the original non-differentiable sample with a differentiable sample. Specifically, $\hat{m}$ is parameterized by a distribution vector $\pi_{l,j} = [\alpha_{l,j}, 1-\alpha_{l,j}]$, where $\alpha_{l,j}$ represents the probability of not sharing the $j$-th parameter in layer $l$ with the server (the probability of mask for the $j$-th parameter in layer $l$ to be zero). $v_{l, j}$ represents the soft \PseudoDefense decision for the $j$-th parameter in layer $l$. The reparameterization trick is used for differentiable training:
\begin{align}
\label{eq:gumbel_soft}
v_{l, j}(k) = \frac{\exp\big((\log\pi_{l,j}(k) + G_{l,j}(k)) / \tau\big)}{\sum\limits_{i \in \{0, 1\}}\exp\big((\log\pi_{l,j}(i) + G_{l,j}(i)) / \tau\big)}
\end{align}

where $G_{l,j}=-\log(-\log U_{l,j})$ is a standard Gumbel distribution with $U_{l,j}$ sampled from a uniform i.i.d. distribution $\text{Unif}(0,1)$ and $\tau$ is the temperature of the softmax. $k \in \{0, 1\}$ and $\alpha_{l,j}$ represents the probability of not sharing the $j$-th parameter in layer $l$ with the server (the probability of mask for the $j$-th parameter in layer $l$ to be zero).

As depicted in Algorithm~\ref{alg:adaptive_PriPrune}, in each FL round, after user $i$ receives the aggregated global model $\bm{w}_{i}^{t}$ from the server, they sample their defense mask $\hat{m}_{i}$ using Eq.(~\ref{eq:gumbel_soft}). This mask $\hat{m}_{i}$ is then employed to combine the global model $\bm{w}^{t}$ and the user's local model $\bm{w}_{i}^{t-1}$. Following this composition step, the user initiates local training based on Eq.(~\ref{eq:loss_func}), leading to updates in $\bm{w}_{i}^{t}$ and $\alpha_{i}^{t}$. Then user $i$ retains the non-shared weights locally and transmits the shared weights $\bm{\hat{w}}_{i}^{t}$ to the server.

Next, we discuss the terms of the loss function in Eq. (\ref{eq:loss_func}) and their rationale. 
First,  the privacy loss $\mathcal{L}_{pri}$ discourages sharing weights with large gradients. Building on Insight 1, we assign higher weights to parameters with larger gradients, thus promoting the sharing of less information with the server. Moreover, we also distribute constraints to layer $i$ based on its parameter count $N_{l}$, in proportion to the total model weights, which ensures uniform layer-wise sparsity. 
In the privacy loss, $\mathcal{L}_{pri}$, the model weights with large gradients are assigned a larger weight in the loss function, thus inducing a larger loss and a higher $\alpha$, and are less likely to be shared with the server. 

\begin{align}
\label{eq:loss_func_pri}
    \mathcal{L}_{pri} = \sum\limits_{l \in L, j \in d_{l}} - \frac{N_{l}}{\sum\limits_{l \in L} N_{l}}\frac{|g_{j}|}{\sum\limits_{j \in d_{l}} |g_{j}|} \log \alpha_{l,j}
\end{align}
where $d_{l}$ denotes the weights in layer $l$ of the model. $N_{l}$ indicates the number of weights in layer $l$, $\alpha_{l,j}$ represents the probability of not sharing the $j$-th parameter in layer $l$ with the server (the probability of the $j$-th parameter in layer $l$ of defense mask $\hat{m}$ being zero).

The accuracy loss  is $\mathcal{L}_{acc} = \mathcal{L} \left ( \mathcal{D}_{i} ; \bm{w}^{t}_{i} \vert \bm{w}^{t}_{i,init}\right)$. 
The initial local weight $\bm{w}^{t}_{i,init} = \bm{w}^{t} \odot \hat{m}^{t-1}_{i} + \bm{w}_{i}^{t-1} \odot {\tilde{m}}^{t-1}_{i} $, 
Where $\tilde{m}$ denotes the bitwise complement matrix of $\hat{m}$. It is the composition of locally preserved weights with global weights through the utilization of the defense mask $\hat{m}$.  Finally,  the term of $\sum\limits \alpha$ in the loss function serves to incorporate a privacy penalty based on the magnitude of $\alpha$.

\subsection{\algoname's Performance} 
We conduct comprehensive experiments and thoroughly evaluate the performance of \algoname for diverse pruning mechanisms and datasets. As mentioned earlier in Insight 2 and 3, we adopt the fixed defense rate of \textit{Largest (Pseudo)} and we illustrate \textit{Largest (Pseudo)} with different defense rates in Figure \ref{fig:priprune_all}. It is worth noting that increasing the defense rate of \textit{Largest (Pseudo)} could achieve more privacy protection but at the cost of less model accuracy. On the other hand, our proposed \algoname dynamically chooses its defense rate to achieve a better balance between privacy and accuracy. When we compare \algoname to \textit{PruneFL}, \algoname not only obtains similar accuracy but also significantly improves privacy, as shown in Fig. \ref{fig:tradeoff_priprune_prunefl}. When compared to \textit{Largest (Pseudo)}, \algoname reaches higher accuracy while still maintaining a similar level of privacy.  Fig. \ref{fig:priprune_prunefl_accuracy} and \ref{fig:priprune_prunefl_NMI} illustrate the performance metrics, accuracy, and privacy with regard to FL rounds. As shown in Figure \ref{fig:priprune_prunefl_accuracy}, the convergence curve of \algoname and its model accuracy closely align with that of PruneFL. Figure~\ref{fig:priprune_prunefl_NMI} illustrates the privacy performance, revealing that \algoname consistently maintains a low normalized MI across all FL training rounds and showing the superiority of \algoname in terms of privacy protection compared to other methods. The adaptive \algoname enables fine-tuning the defense rate based on the model's accuracy through our joint optimization and Figures \ref{fig:priprune_prunefl_accuracy} demonstrate how the defense rate changes in different FL rounds. Figure \ref{fig:priprune_all_methods} demonstrates the effectiveness of \algoname on various pruning methods: Random, SNIP, SynFlow, FedDST, and PruneFL in FEMNIST dataset. Overall, through the adoption of \algoname, all pruning methods achieve enhanced privacy levels without compromising accuracy, which emphasizes the effectiveness and versatility of \algoname.

\paragraph{Hyperparameter Selection}
We conducted hyperparameter searches for our main parameter, the defense rate, from 0.1 to 0.6, as depicted in Fig. 4a. In Insight 3, we incorporate the defense rate into our objective function. The defense rate dynamically adapts, without manual selection, across FL rounds as shown in Fig. 4b. With regards to our hyperparameters in objective functions, we search $\lambda_{acc}$ in the list of [1, 5, 10], $\lambda_{pri}$ in the list of [1, 5, 10, 15, 20] and $\lambda_{sha}$ in the list of [0.0001, 0.00002].

\section{Conclusion}
In this paper, we revisit federated learning with model pruning, from the point of view of privacy. First, we quantify the information leakage and privacy gain offered for model pruning in FL, both theoretically and experimentally. %
Inspired by insights obtained from our extensive evaluation, we design \algoname -- an adaptive privacy-preserving local pruning mechanism in FL, that is jointly optimized for privacy and model performance.  We show the effectiveness of our proposed mechanism across diverse pruning methods and datasets. One direction for future work is  to combine our pruning-based defense with classic, orthogonal defenses in FL such as differential privacy and secure aggregation.

\bibliography{aaai24}

\clearpage
\appendix
\label{sec:appendix}
This appendix serves as a comprehensive extension to the main body of our paper, providing supplementary materials that enhance the understanding of our findings. Divided into two key sections, this appendix first presents the theoretical proofs that underpin our research. Following the theoretical foundation, the appendix delves into additional evaluation details and results, furnishing more pertinent evidence to substantiate our paper.

\subsection{Proof of Theorem}
\label{ap:proof}

\subsubsection{Proof}
The definition of mutual information is: 
\begin{align}\label{eq:mutual_info}
     \mathbf{I} \left (X;Y \right ) 
     & = \int_{\mathcal{X}\times \mathcal{Y}} log \frac{d\mathds{P}_{X,Y}{(x,y)}}{d\mathds{P}_{X}(x)
     \otimes \mathds{P}_{Y}(y)}d\mathds{P}_{X,Y}(x,y)\notag\\
     & = \mathbf{H}(X) -\mathbf{H}\left ( X\middle| Y \right )\notag\\
     & = \mathbf{H}(X) + \mathbf{H}(Y) - \mathbf{H}(X,Y)
\end{align}
Hence, Equation~\ref{eq:multi_round} can be written as:
\begin{align}
\label{eq:multi_to_single}
    &\mathbf{I}\left (  \mathcal{D}_{i} ; \left \{ \bm{w}^{t}_{i} \odot m^{t}_{i}  \right \}_{t \in \left [ T \right ] }\right)\notag\\ 
    & \overset{(a)}{\leq}  \sum_{t=1}^{T} \mathbf{I}\left (  \mathcal{D}_{i} ;  \bm{w}_{i}^{t} \odot m_{i}^{t} \middle| \left \{ \bm{w}^{k}_{i} \odot m^{k}_{i}  \right \}_{k \in \left [ t-1 \right ] }  \right)\notag\\
    & \overset{(b)}{=} \sum_{t=1}^{T} \left(\mathbf{I}\left ( \bm{w}^{t}_{i} ;  \bm{w}_{i}^{t} \odot m_{i}^{t} \middle| \left \{ \bm{w}^{k}_{i} \odot m^{k}_{i}  \right \}_{k \in \left [ t-1 \right ] } \right)\right)\notag\\
    & + \sum_{t=1}^{T} \left(\mathbf{I}\left (  \mathcal{D}_{i} ;  \bm{w}_{i}^{t} \odot m_{i}^{t} \middle| \bm{w}_{i}^{t}, \left \{ \bm{w}^{k}_{i} \odot m^{k}_{i}  \right \}_{k \in \left [ t-1 \right ] } \right)\right)\notag\\
    &-\sum_{t=1}^{T} \left(\mathbf{I}\left ( \mathcal{D}_{i} ; \bm{w}_{i}^{t} \middle| \left \{ \bm{w}^{k}_{i} \odot m^{k}_{i}  \right \}_{k \in \left [ t \right ]} \right) \right) \notag\\
    & \overset{(c)}{\leq} \sum_{t=1}^{T} \left(\mathbf{I} \left ( \bm{w}_{i}^{t} ;  \bm{w}_{i}^{t} \odot m_{i}^{t} \middle| \left \{ \bm{w}^{k}_{i} \odot m^{k}_{i}  \right \}_{k \in \left [ t-1 \right ] } \right)\right)\notag\\
    & =\sum_{t=1}^{T} \left(\mathbf{H} \left ( \bm{w}_{i}^{t}  \middle| \left \{ \bm{w}^{k}_{i} \odot m^{k}_{i}  \right \}_{k \in \left [ t-1 \right ] } \right)\right) \notag\\
    &+ \sum_{t=1}^{T} \left(\mathbf{H} \left (\bm{w}_{i}^{t} \odot m_{i}^{t}  \middle| \left \{ \bm{w}^{k}_{i} \odot m^{k}_{i}  \right \}_{k \in \left [ t-1 \right ] } \right) \right)\notag\\
    & -  \sum_{t=1}^{T} \left(\mathbf{H} \left ( \bm{w}_{i}^{t},\bm{w}_{i}^{t} \odot m_{i}^{t}  \middle| \left \{ \bm{w}^{k}_{i} \odot m^{k}_{i}  \right \}_{k \in \left [ t-1 \right ] } \right)\right) \notag\\
    & \overset{(d)}{=}\sum_{t=1}^{T} \left(\mathbf{I} \left ( \mathrm{x}_{i}^{t} ;  \bm{w}_{i}^{t} \odot m_{i}^{t} \middle| \left \{ \bm{w}^{k}_{i} \odot m^{k}_{i}  \right \}_{k \in \left [ t-1 \right ] } \right)\right)\notag\\
\end{align}
Where (a) and (b) are from the chain rule; (c) is from data processing inequality $\mathcal{D}_{i}\rightarrow \bm{w}_{i}^{t} \rightarrow \bm{w}_{i}^{t} \odot m_{i}^{t} $ that $\mathbf{I}\left (  \mathcal{D}_{i} ; \bm{w}_{i}^{t} \odot m_{i}^{t} \middle| \bm{w}_{i}^{t}, \left \{ \bm{w}^{k}_{i} \odot m^{k}_{i}  \right \}_{k \in \left [ t-1 \right ] } \right) = 0$; (d) is from zero conditional entropy and invariant of mutual information, where 
\begin{align}
    \mathrm{x}_{i}^{t} = \frac{\partial \ell_{i}({\hat{\bm{w}}_{i}^{t-1};\mathcal{D}_{i}})}{\partial \bm{w}}=  \frac{1}{B}\sum_{b \in \mathbf{B}_{i}^{t}}g_{i}\left (\bm{w}_{i}^{t},b\right )
\end{align}
$b$ denotes the size of the random samples.

Therefore, the privacy leakage for a single round $t$ is
\begin{align}
\label{eq:single_round}
    \mathbf{I}\left ( \mathrm{x}_{i}^{t} ;  \bm{w}_{i}^{t} \odot m_{i}^{t} \middle| \left \{ \bm{w}^{k}_{i} \odot m^{k}_{i}  \right \}_{k \in \left [ t-1 \right ] }   \right ) 
\end{align}
We first prove the upper bound for Equation~\ref{eq:single_round}.

Based on Equation~\ref{eq:mutual_info}, we have
\begin{align}
     & \mathbf{I}\left ( \mathrm{x}_{i}^{t} ;  \bm{w}_{i}^{t} \odot m_{i}^{t} \middle| \left \{ \bm{w}^{k}_{i} \odot m^{k}_{i}  \right \}_{k \in \left [ t-1 \right ] }   \right ) \notag\\
      & = \mathbf{H}\left ( \mathrm{x}_{i}^{t} \middle| \left \{ \bm{w}^{k}_{i} \odot m^{k}_{i}  \right \}_{k \in \left [ t-1 \right ] } \right )  + \mathbf{H}\left ( \bm{w}_{i}^{t} \odot m_{i}^{t} \middle| \left \{ \bm{w}^{k}_{i} \odot m^{k}_{i}  \right \}_{k \in \left [ t-1 \right ] }   \right ) \notag\\
      & - \mathbf{H}\left ( \mathrm{x}_{i}^{t} , \bm{w}_{i}^{t} \odot m_{i}^{t} \middle| \left \{ \bm{w}^{k}_{i} \odot m^{k}_{i}  \right \}_{k \in \left [ t-1 \right ] }   \right ) \notag\\
      & \leq \underset{\mathbf{A}}{\underbrace{\mathbf{H}\left ( \bm{w}_{i}^{t} \odot m_{i}^{t} \middle| \left \{ \bm{w}^{k}_{i} \odot m^{k}_{i}  \right \}_{k \in \left [ t-1 \right ] }   \right )}}+ \underset{\mathbf{B}}{\underbrace{\mathbf{H}\left ( \mathrm{x}_{i}^{t} \middle| \left \{ \bm{w}^{k}_{i} \odot m^{k}_{i}  \right \}_{k \in \left [ t-1 \right ] } \right)}} \notag\\
\end{align}

For part $\mathbf{A}$, based on the chain rule of entropy, we have
\begin{align}
  0 \leq & \mathbf{H}\left (\bm{w}_{i}^{t},m_{i}^{t}, \bm{w}_{i}^{t} \odot m_{i}^{t} \middle| \left \{ \bm{w}^{k}_{i} \odot m^{k}_{i}  \right \}_{k \in \left [ t-1 \right ] }\right) \notag\\
 & =  \mathbf{H}\left (\bm{w}_{i}^{t},m_{i}^{t} \middle| \left \{ \bm{w}^{k}_{i} \odot m^{k}_{i}  \right \}_{k \in \left [ t-1 \right ] }\right) \notag\\
 & +\mathbf{H}\left ( \bm{w}_{i}^{t} \odot m_{i}^{t} \middle| \bm{w}_{i}^{t},m_{i}^{t} , \left \{ \bm{w}^{k}_{i} \odot m^{k}_{i}  \right \}_{k \in \left [ t-1 \right ] }\right)\notag\\
 & =  \mathbf{H}\left (\bm{w}_{i}^{t} \odot m_{i}^{t} \middle| \left \{ \bm{w}^{k}_{i} \odot m^{k}_{i}  \right \}_{k \in \left [ t-1 \right ] }\right) \notag\\
 & +\mathbf{H}\left ( \bm{w}_{i}^{t},m_{i}^{t} \middle| \left \{ \bm{w}^{k}_{i} \odot m^{k}_{i}  \right \}_{k \in \left [ t \right ] }\right)\notag\\
\end{align}
Due to $\mathbf{H}\left ( \bm{w}_{i}^{t} \odot m_{i}^{t} \middle| \bm{w}_{i}^{t},m_{i}^{t} , \left \{ \bm{w}^{k}_{i} \odot m^{k}_{i}  \right \}_{k \in \left [ t-1 \right ] }\right) = 0$, and $\mathbf{H}\left ( \bm{w}_{i}^{t},m_{i}^{t} \middle| \left \{ \bm{w}^{k}_{i} \odot m^{k}_{i}  \right \}_{k \in \left [ t \right ] }\right)\geq 0$, we have
\begin{align}
    &\mathbf{H}\left (\bm{w}_{i}^{t} \odot m_{i}^{t} \middle| \left \{ \bm{w}^{k}_{i} \odot m^{k}_{i}  \right \}_{k \in \left [ t-1 \right ] }\right)\notag\\ 
    &\leq \mathbf{H}\left (\bm{w}_{i}^{t},m_{i}^{t} \middle| \left \{ \bm{w}^{k}_{i} \odot m^{k}_{i}  \right \}_{k \in \left [ t-1 \right ] }\right) \notag\\
    & \overset{(a)}{\leq}\mathbf{H}\left (\bm{w}_{i}^{t} \middle| \left \{ \bm{w}^{k}_{i} \odot m^{k}_{i}  \right \}_{k \in \left [ t-1 \right ] }\right) + \mathbf{H}\left (m_{i}^{t} \middle| \left \{ \bm{w}^{k}_{i} \odot m^{k}_{i}  \right \}_{k \in \left [ t-1 \right ] }\right) \notag\\
    & \overset{(b)}{=}\mathbf{H}\left ( \mathrm{x}_{i}^{t} \middle| \left \{ \bm{w}^{k}_{i} \odot m^{k}_{i}  \right \}_{k \in \left [ t-1 \right ] } \right) + 1- \frac{1}{2\ln2}\sum^{\infty}_{n=1}\frac{(2p_{i}-1)^{2n}}{n(2n-1)}\notag\\
     & \leq \underset{\mathbf{B}}{\underbrace{\mathbf{H}\left ( \mathrm{x}_{i}^{t} \middle| \left \{ \bm{w}^{k}_{i} \odot m^{k}_{i}  \right \}_{k \in \left [ t-1 \right ] } \right)}} + 1 - \frac{p_{i}-1}{2\ln2}
\end{align}
(a) is from conditioning reduces entropy; (b) is from zero conditional entropy and the Taylor series of the binary entropy function in a neighborhood of 0.5 with the base pruning rate $p_{i}$.

For part $\mathbf{B}$,

\begin{align}
\label{eq:batch_sum_gradient}
   & \mathbf{H}\left ( \mathrm{x}_{i}^{t} \middle| \left \{ \bm{w}^{k}_{i} \odot m^{k}_{i}  \right \}_{k \in \left [ t-1 \right ] } \right)\notag\\
   & = \mathbf{H}\left ( \frac{1}{B}\sum_{b \in \mathbf{B}_{i}^{t}}g_{i}\left (\bm{w}_{i}^{t},b\right ) \middle| \left \{ \bm{w}^{k}_{i} \odot m^{k}_{i}  \right \}_{k \in \left [ t-1 \right ] } \right)
\end{align}
Let $X =\sum_{b \in \mathbf{B}_{i}^{t}}g_{i}\left (\bm{w}_{i}^{t},b\right )$, $Z=\left \{ \bm{w}^{k}_{i} \odot m^{k}_{i}  \right \}_{k \in \left [ t-1 \right ] }$, $Y=\frac{1}{B}X$. $f_{X,Z}(x,z)$ and $f_{Y,Z}(y,z)$ are the corresponding joint probability density functions. 

Then Equation~\ref{eq:batch_sum_gradient} can be written as
\begin{align}
   \mathbf{H}\left ( Y\middle| Z \right) & = -\int_{\mathcal{Y},\mathcal{Z}} f_{Y, Z}\left(y,z \right) \log \frac{f_{Y,Z}(y,z)}{f_{Z}(z)} \dd{y} \dd{z} \notag\\
   & = -\int_{\mathcal{X},\mathcal{Z}} \left|B\right| f_{X,Z}\left(By,z \right) \log \left|B\right|\frac{f_{X,Z}(By,z)}{f_{Z}(z)} \dd{y} \dd{z} \notag\\
   & = -\int_{\mathcal{X},\mathcal{Z}} f_{X,Z}\left(x,z \right) \log\frac{f_{X,Z}(x,z)}{f_{Z}(z)} \dd{x} \dd{z} + log\frac{1}{\left|B\right|}\notag\\
   & =  \underset{\mathbf{C}}{\underbrace{\mathbf{H}\left ( \sum_{b \in \mathbf{B}_{i}^{t}}g_{i}\left (\bm{w}_{i}^{t},b\right ) \middle| \left \{ \bm{w}^{k}_{i} \odot m^{k}_{i}  \right \}_{k \in \left [ t-1 \right ] } \right)}} + \log\frac{1}{B}
\end{align}
In recent theoretical results for analyzing the behavior of SGD, they approximate the SGD vector by a distribution with independent components or by a multivariate Gaussian vector. 

We define $\hat{g_{i}}\left (\bm{w}_{i}^{t},b\right) \in \mathds{R}^{d^{*}}$ is the largest sub-vector of the $g_{i}\left (\bm{w}_{i}^{t},b\right)$ with non-singular convariance matrices, where $d^{*}\leq d$. Based on the ZCA whitening transformation and Assumption~\ref{as:guassian}, we have $ \hat{g_{i}}\left (\bm{w}_{i}^{t},b\right) = \Sigma_{i}^{-\frac{1}{2}}\bm{z}$, where $\bm{z}$ has zero mean and $\mathds{I}_{d^{*}}$ convariance matrix.

Then for the part $\mathbf{C}$, we set $\bm{v} = \sum_{b \in \mathbf{B}_{i}^{t}}\bm{z}$ and $ \bm{u} = \Sigma_{i}^{-\frac{1}{2}}\bm{v}$, then we have:
\begin{align}
\label{eq:part_C}
    \mathbf{C} & = \mathbf{H}\left ( \sum_{b \in \mathbf{B}_{i}^{t}}\hat{g_{i}}\left (\bm{w}_{i}^{t},b\right ) \middle| \left \{ \bm{w}^{k}_{i} \odot m^{k}_{i}  \right \}_{k \in \left [ t-1 \right ] } \right)\notag\\
   & = \mathbf{H}\left (\Sigma_{i}^{-\frac{1}{2}} \sum_{b \in \mathbf{B}_{i}^{t}}\bm{z} \middle| \left \{ \bm{w}^{k}_{i} \odot m^{k}_{i}  \right \}_{k \in \left [ t-1 \right ] } \right)\notag\\
   & = -\int\mathds{P}_{\mathcal{U}}(\bm{u})\log \mathds{P}_{\mathcal{U}}(\bm{u}) \dd{\bm{u}}\notag\\
   & \overset{(a)}{=} -\int\frac{\mathds{P}_{\mathcal{V}}(\bm{v})}{\left|\det(\Sigma_{i}^{-\frac{1}{2}})\right|}\log\left(\frac{\mathds{P}_{\mathcal{V}}(\bm{v})}{\left|\det(\Sigma_{i}^{-\frac{1}{2}})\right|} \right)\left|\det(\Sigma_{i}^{-\frac{1}{2}})\right|\dd{\bm{v}}\notag\\
   & = \log\left|\det(\Sigma_{i}^{-\frac{1}{2}})\right|+ \underset{\mathbf{D}}{\underbrace{\mathbf{H}\left (\sum_{b \in \mathbf{B}_{i}^{t}}\bm{z} \middle| \left \{ \bm{w}^{k}_{i} \odot m^{k}_{i}  \right \}_{k \in \left [ t-1 \right ] } \right)}}
\end{align}
(a) is based on transformation of random vectors.
Based on the Maximum Entropy Upper Bound~\cite{thomas2006elements}, which the continuous distribution $\mathrm{X}$ with prescribed variance $\mathcal{V}(\mathrm{X})$ maximizing the entropy is the Gaussian distribution of same variance. 
Since the entropy of a multivariate Gaussian distribution with mean $\mu$ and covariance $K_{i,j}$ is
\begin{align*}
    \mathbf{h}(\mathcal{N}_n(\mu,K)) = \frac{1}{2}\log \left (2 \pi e\right) ^{n}\left | K \right |
\end{align*}
where $\left | K \right |$ denotes the determinant of $K$.
Hence, the maximum entropy upper bound for part $\mathbf{D}$ is
\begin{align}
\label{eq:part_D}
    \mathbf{H}\left (\sum_{b \in \mathbf{B}_{i}^{t}}\bm{z} \middle| \left \{ \bm{w}^{k}_{i} \odot m^{k}_{i}  \right \}_{k \in \left [ t-1 \right ] } \right) 
    & \leq \frac{1}{2}\log \left (2 \pi e\right) ^{d^{*}}\left | \mathds{I}_{d^{*}} \right |\notag\\
    & = \frac{d^{*}}{2}\log \left (2 \pi e\right)
\end{align}

Therefore, by combining Equation~\ref{eq:part_C} and Equation~\ref{eq:part_D}, the upper bound for part $\mathbf{B}$ is 
\begin{align}
    \mathbf{H}\left ( \mathrm{x}_{i}^{t} \middle| \left \{ \bm{w}^{k}_{i} \odot m^{k}_{i}  \right \}_{k \in \left [ t-1 \right ] } \right)  \leq \log\frac{1}{B} & +  \log\left|\det(\Sigma_{i}^{-\frac{1}{2}})\right| \notag\\
    & + \frac{d^{*}}{2}\log \left (2 \pi e\right) \notag\\
\end{align}

After summing part $\mathbf{A}$ and part $\mathbf{B}$, we derive the upper bound for the privacy leakage for the single round, which is Equation~\ref{eq:single_round}:
\begin{align}
    &\mathbf{I}\left ( \mathrm{x}_{i}^{t} ;  \bm{w}_{i}^{t} \odot m_{i}^{t} \middle| \left \{ \bm{w}^{k}_{i} \odot m^{k}_{i}  \right \}_{k \in \left [ t-1 \right ] }   \right ) \notag\\
    & \leq 1 - \frac{p_{i}-1}{2\ln2} + 2\log\frac{1}{B}
     +  2\log\left|\det(\Sigma_{i}^{-\frac{1}{2}})\right| 
    + d^{*}\log \left (2 \pi e\right) \notag\\
\end{align}

Based on Equation~\ref{eq:multi_to_single}, we have the upper bound for objective function~\ref{eq:multi_round}, standing how much information the aggregated model over $T$ global training rounds could leak about the private data,
\begin{align}
     & \mathbf{I}_{i} \leq T\left( 1 - \frac{p_{i}-1}{2\ln2} + 2\log\frac{1}{B}
     +  2\log\left|\det(\Sigma_{i}^{-\frac{1}{2}})\right| 
    + d^{*}\log \left (2 \pi e\right)\right)
\end{align}

For PruneFL, the upper bound for the single-round leakage is
\begin{align}
     \mathbf{I}_{i}^{t} & \leq 1 + \frac{\bar{\mathcal{P}}\cup\mathcal{A}}{2d\ln2} + 2\log\frac{1}{B}
     +  2\log\left|\det(\Sigma_{i}^{-\frac{1}{2}})\right| 
    + d^{*}\log \left (2 \pi e\right)\notag\\
     & \leq 1 + \frac{\bar{\mathcal{P}}}{2d\ln2} + 2\log\frac{1}{B}
     +  2\log\left|\det(\Sigma_{i}^{-\frac{1}{2}})\right| 
    + d^{*}\log \left (2 \pi e\right)
\end{align}

\subsection{Privacy Attacks}

\subsubsection{Extra Implementation details} Our algorithms are implemented by Pytorch and we implement experiments on two NVIDIA RTX A5000 and two Xeon Silver 4316. The $\lambda_{acc}$ is selected from [1, 5, 10],  $\lambda_{pri}$ is selected from [1, 10, 15] and $\lambda_{sha}$ is selected from [$2.0\times {{10}^{-5}}$, $2.0\times {{10}^{-6}}$]. Each user will do FedSGD locally and the total FL rounds is 20000. Across all the examined methods, we initialize the learning rate to 0.25 and The baseline pruning rate is set at 0.3 for all datasets.

For the FEMNIST dataset, the basic batch size is 20, and each local update is performed once. 
On the other hand, for the CIFAR10 dataset, the basic batch size is 1, and the local update is executed once as well.

\subsubsection{Comparison of Attacks}
\begin{figure}[htbp]
    \centering
    \includegraphics[width=\columnwidth]{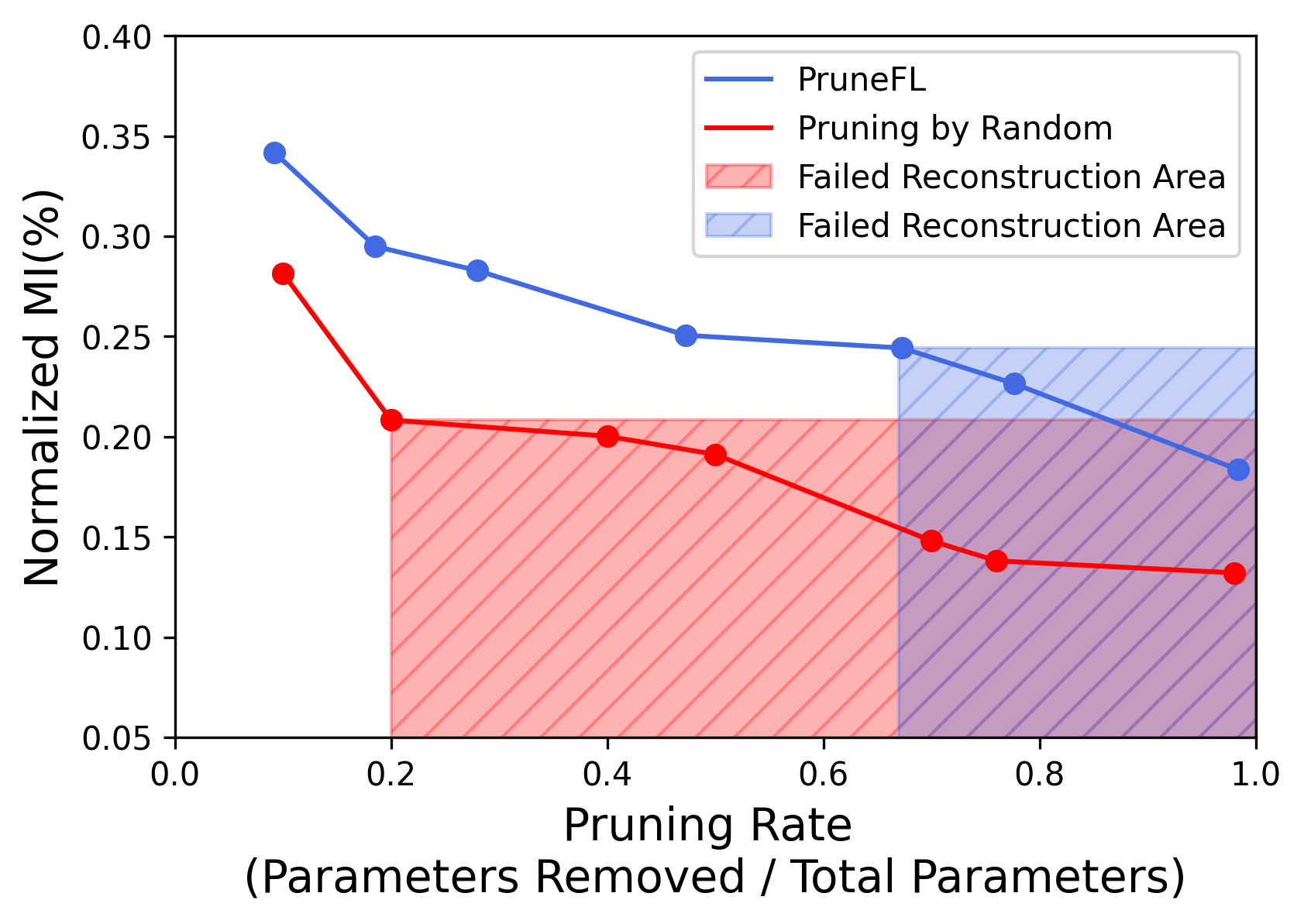}
    \caption{Impact of Varying Pruning Rates on Reconstructed Image Quality Using Normalized Mutual Information (NMI) Metric in DLG Attack.
 }
    \label{fig:Comparison_NMI}
\end{figure}

In this subsection, we initiate our attack analysis by employing the classic Gradient Inversion (GI) attack~\cite{geiping2020inverting}, a member of the DLG attack family, applied to both the Random and PruneFL pruning methods. As depicted in Figure~\ref{fig:Comparison_NMI}, under this classic DLG attack, the reconstruction of local image data fails even with low pruning rates, such as $p=0.2$ for Random pruning.

This could be attributed to the sparsity patterns introduced by model pruning in FL.  Our exploration takes a step further to optimize the attack strategy. The goal is to capitalize on the characteristics of sparsity inherent to the model pruning context. Consequently, we introduce an advanced attack tailored for the context of model pruning within FL, termed the Sparse Gradient Inversion (SGI) attack. This novel approach aims to recover the pruning mask embedded within the pruned model.

\begin{figure}[!htbp]
    \centering
    \includegraphics[width=\columnwidth]{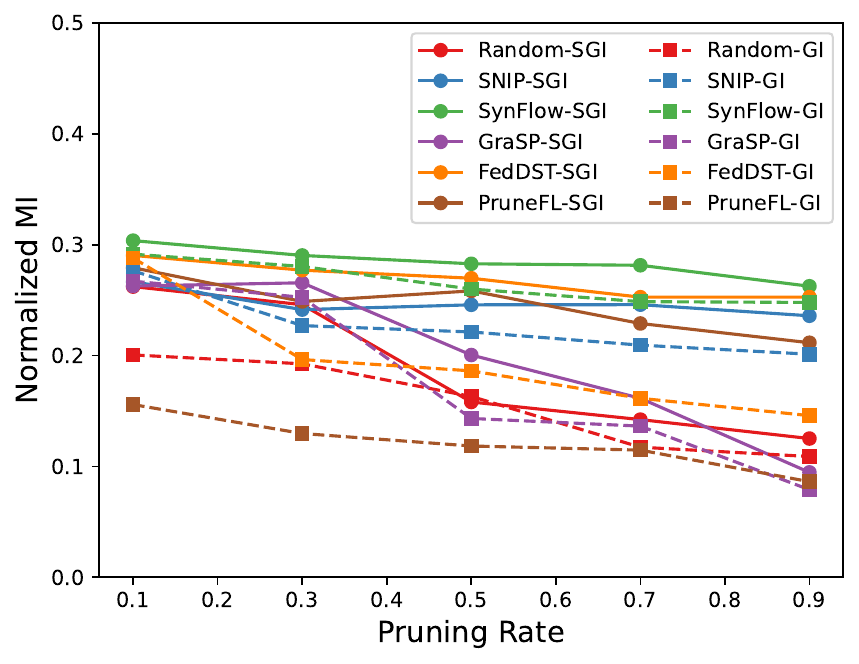}
    \caption{Comparison between the SGI (Sparse Gradient Inversion) attack and the Gradient Inversion attack on FEMNIST dataset  for six pruning methods under varying the base pruning rate.}
    \label{fig:SGI_GI}
\end{figure}

We compare the effectiveness of the Sparse Gradient Inversion (SGI) attack with the Gradient Inversion attack under varying the base pruning rate. Our evaluation is based on the normalized mutual information (NMI) metric. Specifically, we analyze the NMI achieved by each attack for different pruning methods. Our findings in Figure~\ref{fig:SGI_GI} demonstrate that the SGI attack consistently outperforms the Gradient Inversion attack consistently 
by inferring a larger amount of data at each pruning rate level, across all tested pruning methods. This highlights the efficacy of the SGI attack in exploiting the sparsity inherent to model pruning, ultimately leading to more pronounced privacy breaches.

\subsubsection{Evaluation on both NMI and PSNR Metrics}
\begin{figure}[!htbp]
     \centering
     \includegraphics[width=\linewidth]{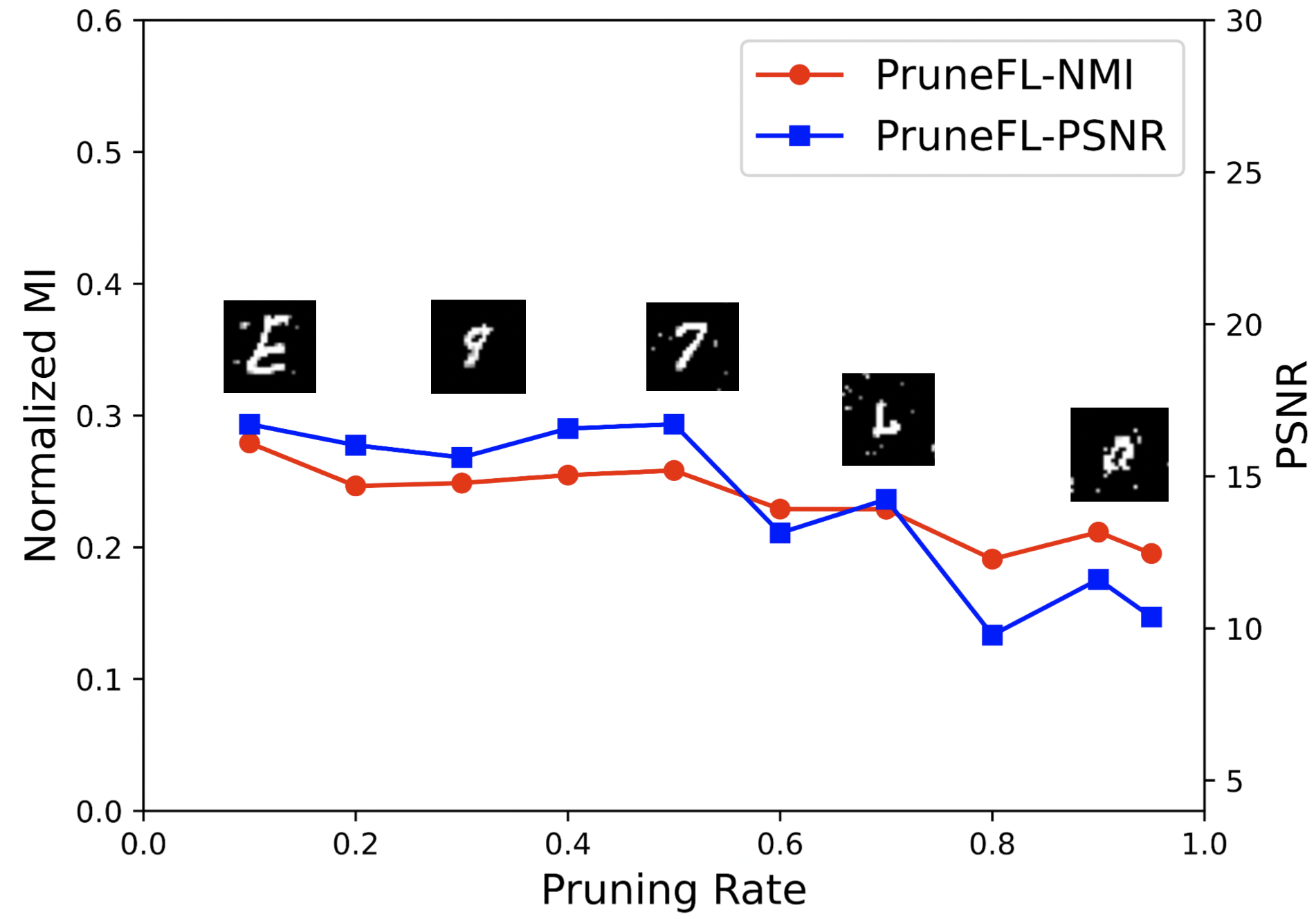}
      \caption{Impact of varying pruning rate in FEMNIST on privacy leakage using key metrics: Normalized Mutual Information (NMI) displayed on the left y-axis and  Peak Signal-to-Noise Ratio (PSNR) shown on the right y-axis  under a Gradient Inversion attack for PruneFL.
   }
\label{fig:pruning_rate_prunefl}
\end{figure}

To quantify the extent of privacy leakage, we utilized the Normalized Mutual Information (NMI) metric, which has been demonstrated in prior research to align well with the Peak Signal-to-Noise Ratio (PSNR) as an effective measure of privacy leakage.
The experimental result, as depicted in Figure~\ref{fig:pruning_rate_prunefl}, reveals that even with a high pruning rate of 0.9, the PruneFL exhibits inadequate defense against privacy attacks. This indicates that the pruned model retains vulnerabilities against fundamental privacy attacks. allowing for potential information leakage and divulging sensitive details about individual users' local datasets during the FL process.  
Furthermore, the observed relationship between higher pruning rate and reduced Normalized Mutual Information (NMI) and the Peak Signal-to-Noise Ratio (PSNR) metrics reaffirms and aligns with previous research findings~\cite{elkordy2023much}. This corroborates the effectiveness of NMI as a reliable measure of privacy leakage, similar to the well-established PSNR metric.

\begin{figure}[!htbp]
    \centering
    \centering
    \includegraphics[width=\linewidth]{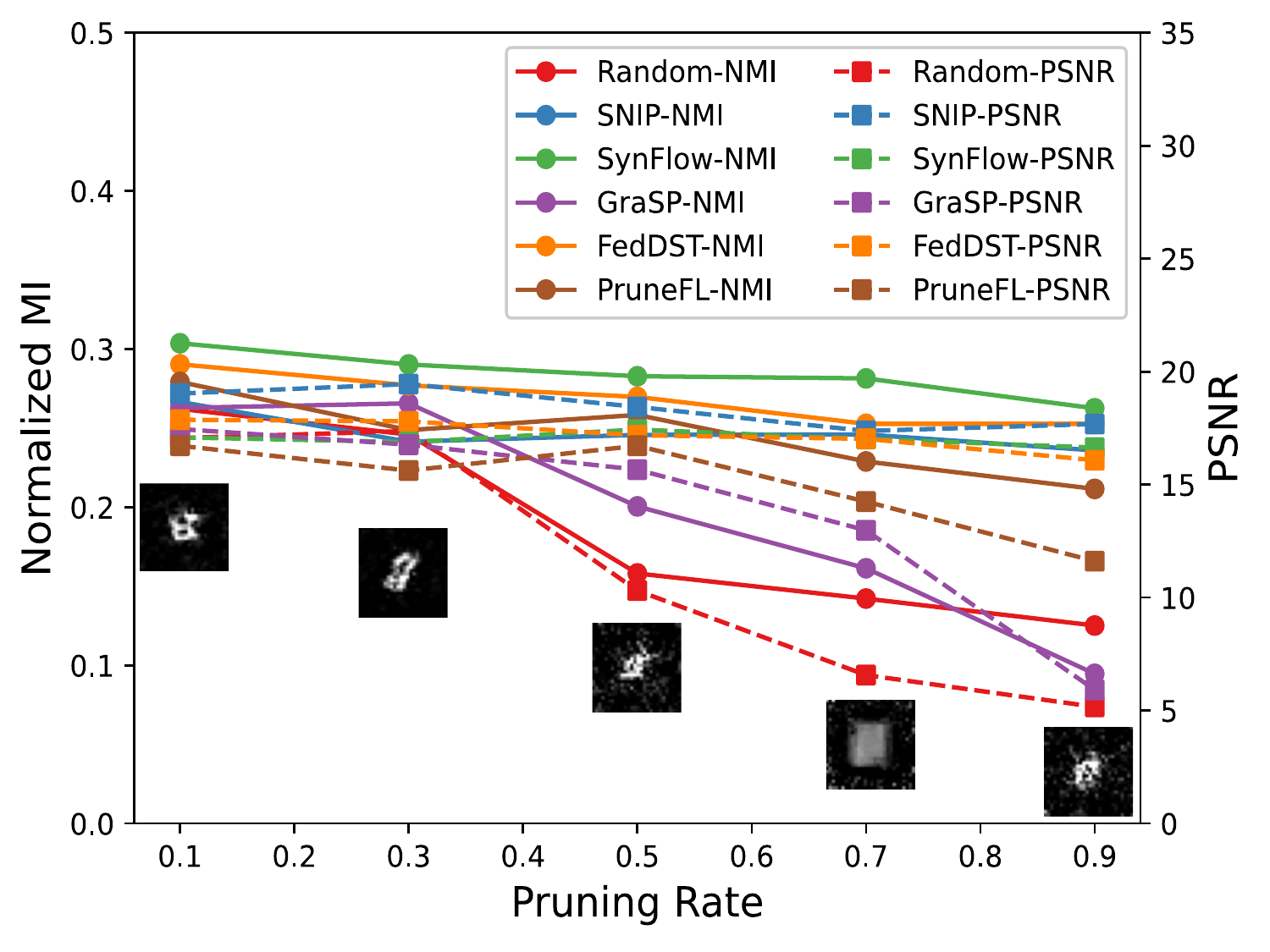}
    \caption{Impact of varying the base pruning rate in FEMNIST on privacy leakage using key metrics: Normalized Mutual Information (MI) displayed on the left y-axis and  Peak Signal-to-Noise Ratio (PSNR) shown on the right y-axis under a Sparse Gradient Inversion attack for six pruning methods: Random, SNIP, SynFlow, GraSP, FedDST, and PruneFL.}\label{fig:pruning_rate_femnist_psnr}
\end{figure}

Then we value the impact of varying pruning rates $p$ selected from [0.1, 0.3, 0.5, 0.7, 0.9] on privacy leakage using NMI and PSNR for six pruning methods: Random, SNIP, SynFlow, GraSP, FedDST, and PruneFL in FEMNIST dataset. As illustrated in Figure~\ref{fig:pruning_rate_femnist_psnr}, the relationship between higher pruning rates and the decrease in NMI and PSNR metrics supports earlier findings. This further validates the effectiveness of NMI as a consistent measure of privacy leakage as the PSNR metric.

\subsection{Evaluation Methods}
In this subsection, we present a overview of the pruning criteria, cycles, and schedules employed by the six evaluated methods. These specifics are elaborated in Table~\ref{tab:evaluated_methods}. By scrutinizing the table, we discern notable trends: three of the methods execute pruning exclusively at the server side, while the remaining methods engage in pruning activities at both the server and client sides. The latter approaches adopt an iterative pruning strategy throughout the course of the FL process.
\begin{table*}[!htbp]
\centering
\begin{tabular}{lclcc}
\hline
\multicolumn{1}{c}{Method} &
  \begin{tabular}[c]{@{}c@{}}Pruning  Criteria\end{tabular} &
  \multicolumn{1}{c}{\begin{tabular}[c]{@{}c@{}}Pruning\\ Cycles\end{tabular}} &
  \begin{tabular}[c]{@{}c@{}}Pruning\\ at Server\end{tabular} &
  \begin{tabular}[c]{@{}c@{}}Pruning\\ at Client\end{tabular} \\ \hline
SNIP~\cite{lee2018snip}    & gradient-based  & one shot  & \checkmark & \ding{55} \\
SynFlow~\cite{tanaka2020synflow} & gradient-based  & one shot  & \checkmark & \ding{55} \\
GraSP~\cite{wang2020grasp}   & gradient-based  & one shot  & \checkmark & \ding{55} \\
Random Pruning  & magnitude-based & iterative & \checkmark & \checkmark  \\
FedDST~\cite{bibikar2022feddst}  & magnitude-based & iterative & \checkmark & \checkmark  \\
PruneFL~\cite{jiang2022pruneFL} & magnitude-based & iterative & \checkmark & \checkmark  \\ \hline
\end{tabular}
\caption{Comparison for Evaluated Methods}
\label{tab:evaluated_methods}
\end{table*}

\subsubsection{Attack Performance on CIFAR10 Dataset}
\begin{figure*}[htbp]
  \centering
  \begin{subfigure}[b]{0.23\textwidth}
    \centering
    \includegraphics[width=\linewidth]{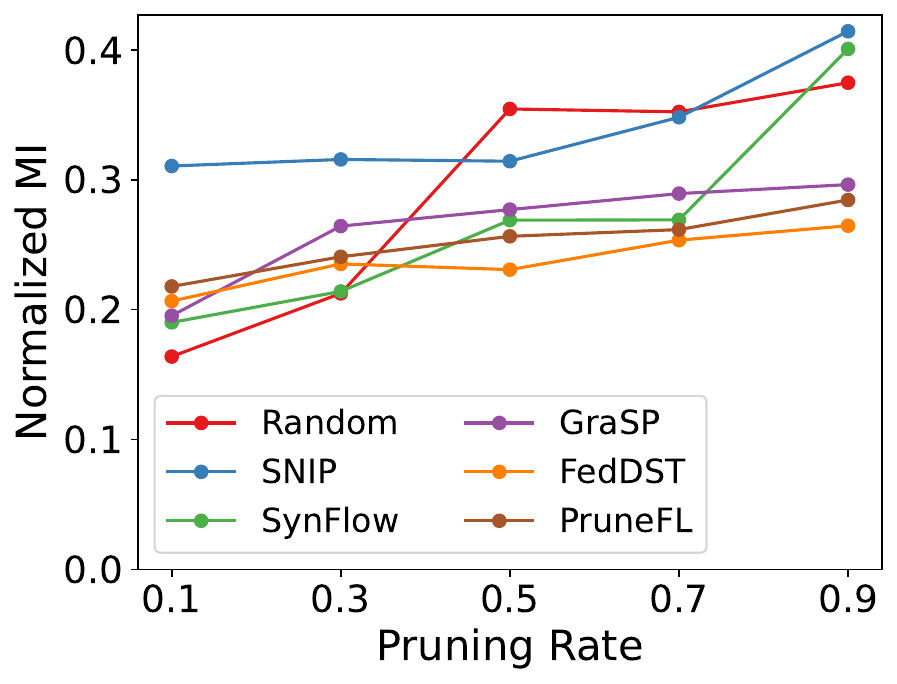}
    \caption{Impact of varying the pruning rate in CIFAR10.}\label{fig:pruning_rate_cifar}
  \end{subfigure}
  \hfill
  \begin{subfigure}[b]{0.23\textwidth}
    \centering
   \includegraphics[width=\linewidth]{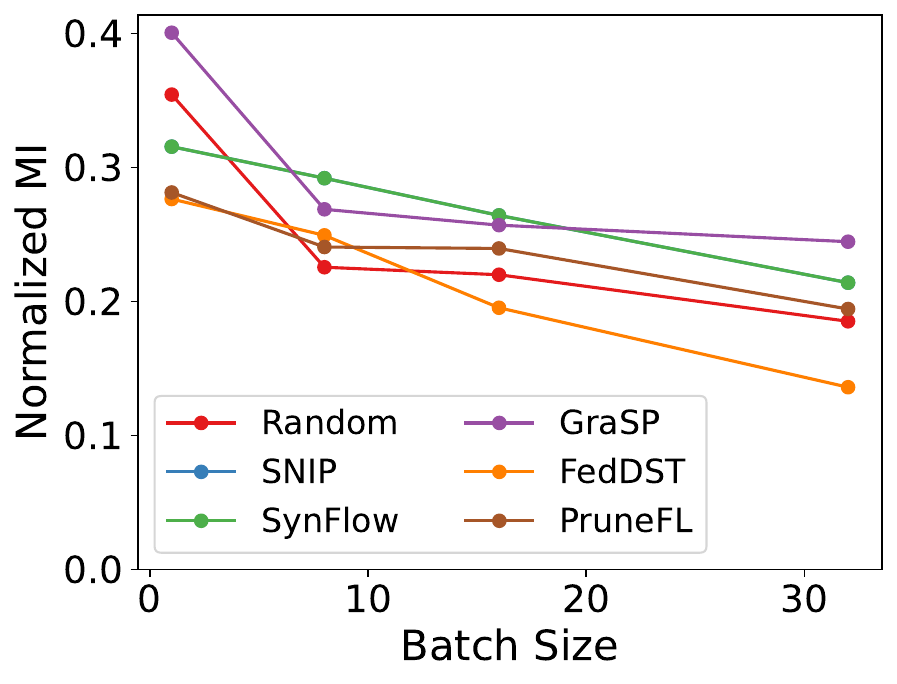}
    \caption{Impact of varying the batch size in CIFAR10.
    }
\label{fig:batch_size_cifar}
  \end{subfigure}
  \hfill
  \begin{subfigure}[b]{0.23\textwidth}
    \centering
    \includegraphics[width=\linewidth]{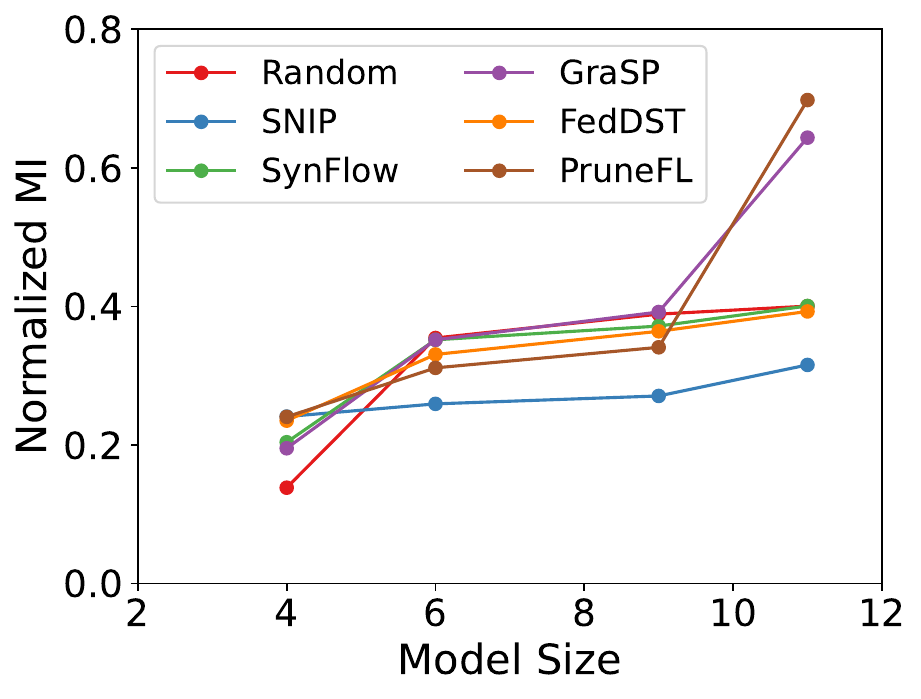}
    \caption{Impact of varying the model size $d$ in CIFAR10.
    }
\label{fig:model_size_cifar}
  \end{subfigure}
  \hfill
  \begin{subfigure}[b]{0.23\textwidth}
    \centering
    \includegraphics[width=\linewidth]{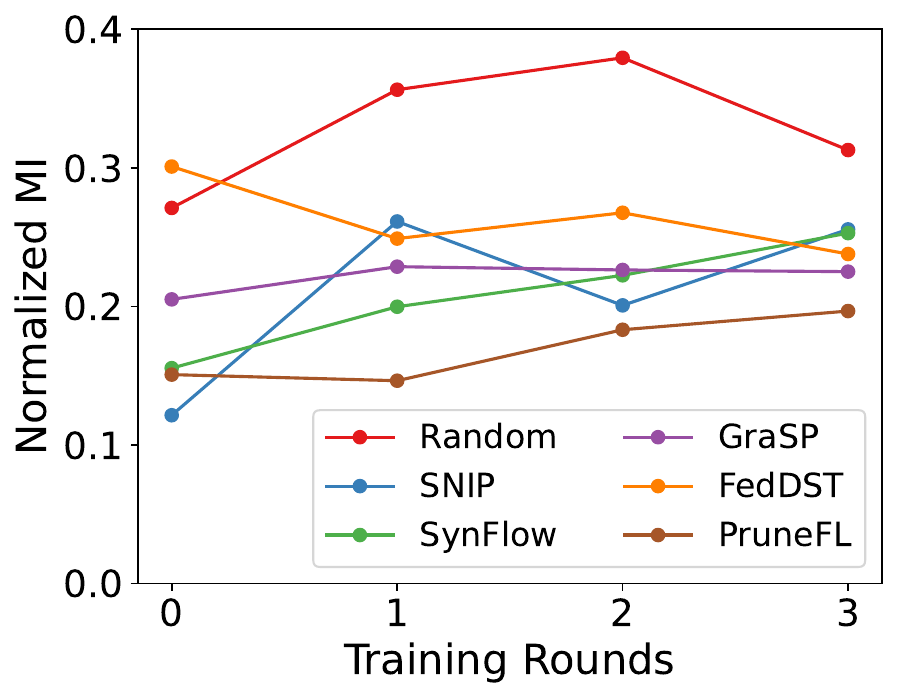}
    \caption{Impact of varying the training round $T$ in CIFAR10.
    }
\label{fig:training_round_cifar}
  \end{subfigure}
 \caption{Impact of varying pruning rate, batch size, model size, and training rounds on privacy leakage using Normalized Mutual Information (NMI) under a Sparse Gradient Inversion (SGI) attack for six pruning methods: Random, SNIP, SynFlow, GraSP, FedDST, and PruneFL.}\label{fig:attack_all_methods_cifar}
\end{figure*}

Considering space constraints, we present the attack performance outcomes for the CIFAR10 dataset in this section. Illustrated in Figure~\ref{fig:attack_all_methods_cifar} it shows the impact of varying pruning rate, batch size, model size, and training rounds on privacy leakage using Normalized
Mutual Information (NMI) under a Sparse Gradient Inversion (SGI) attack for six pruning methods: Random, SNIP, SynFlow,
GraSP, FedDST, and PruneFL, on CIFAR10 dataset.

\noindent\textbf{Impact of Pruning Rate ($p$)}
Figure~\ref{fig:pruning_rate_cifar} illustrates the impact of varying pruning rates on privacy leakage using NMI. By controlling the pruning rate from {0.1,0.3,0.5,0.7,0.9}, we examine the trade-off between privacy leakage and the extent of model pruning. Consistent with our Theory~\ref{th:single_round}, we observe that higher pruning rates lead to a proportionally decreasing trend in NMI. This decrease in NMI indicates a diminished risk of privacy leakage due to the reduced amount of information available in the pruned model.

\noindent\textbf{Impact of Batch Size ($B$)}
Figure~\ref{fig:batch_size_cifar} shows the impact of varying batch size $B$ on privacy leakage using NMI on the CIFAR10 dataset, where $B$ is selected from the set {1, 8, 16, 32}.
It demonstrates that larger batch sizes increase the privacy protection of the FL process as they enable more data points to be added during local training, which can help in obscuring the data characteristics of a particular user.

\noindent\textbf{Impact of Model Size ($d$)}
We consider 4 different model architectures: VGG-4, VGG-6, VGG-9, and VGG-11. These models consist of 1 convolutional layer and 2 linear layers, 4 convolutional layers and 2 linear layers, 6 convolutional layers with 3 linear layers, and 8 convolutional layers with 3 linear layers, respectively. These models encompass 1,125,642, 1,814,538, 6,436,106, and 9,385,994 parameters, respectively.
Figure~\ref{fig:model_size_cifar} depicts the relationship between the model size ($d$) and the corresponding increase in information leakage. The x-axis in the figure represents the number of layers in each model. This observation can be attributed to the fact that larger models possess a higher number of parameters, thereby amplifying gradient information leakage.

\noindent\textbf{Impact of Communication Round ($T$)}Figure~\ref{fig:training_round_cifar} illustrates the impact of training rounds $T$ on  information leakage. Generally, an increase in training rounds results in more information leakage, but it's worth noting that the increase is not linear for all rounds. This observation suggests that the upper bound provided by Corollary~\ref{co:multi_round} for $T$ is not tightly bound.

\subsection{Defense Details}
Within the scope of the three defense methods in the Insights from Evaluation Section, specific configurations have been established. These settings are detailed below:

For all three defense methods, the total pruning rate ($\hat{p}$) is uniformly set at 0.3, mirroring the original PruneFL's pruning rate.

For each defense method:
\begin{itemize}
    \item  PruneFL + Largest: The defense strategy involves pruning based on the weights with top-$k$ largest gradients, with the defense pruning rate set to $\hat{p}_{largest}$ = 0.3.
    \item  PruneFL + Random: The defense strategy involves random pruning, with the defense pruning rate ($\hat{p}$) set to $\hat{p}_{random}$ = 0.3.
    \item PruneFL + Mix: This is a hybrid approach that combines both the largest gradient-based pruning and random pruning. The defense pruning rate is determined as the sum of $\hat{p}_{largest}$ = 0.15 and $\hat{p}_{random}$ = 0.15.
\end{itemize}

Additionally, we explore various combinations of the pruning rates for largest and random within the Mix strategy, including $0\%$ largest with $30\%$ random, $10\%$ largest with $20\%$ random, $20\%$ largest with $10\%$ random, and $30\%$ largest with $0\%$ random.

\begin{figure*}[!htbp]
  \centering
  \begin{subfigure}[b]{0.3\textwidth}
    \includegraphics[width=\linewidth]{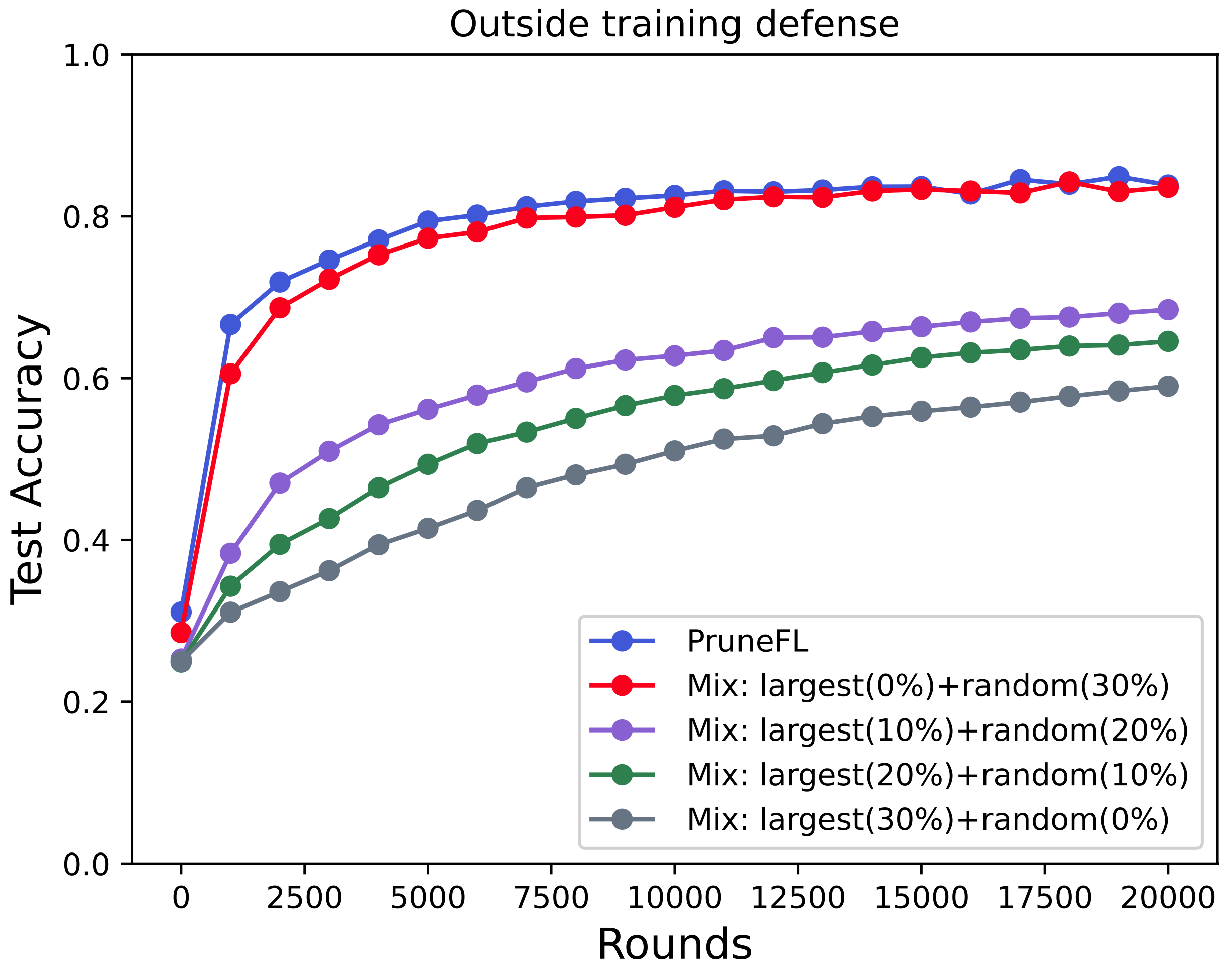}
    \caption{Accuracy Performance}
    \label{fig:acc_mix}
  \end{subfigure}
  \hfill
  \begin{subfigure}[b]{0.3\textwidth}
    \includegraphics[width=\linewidth]{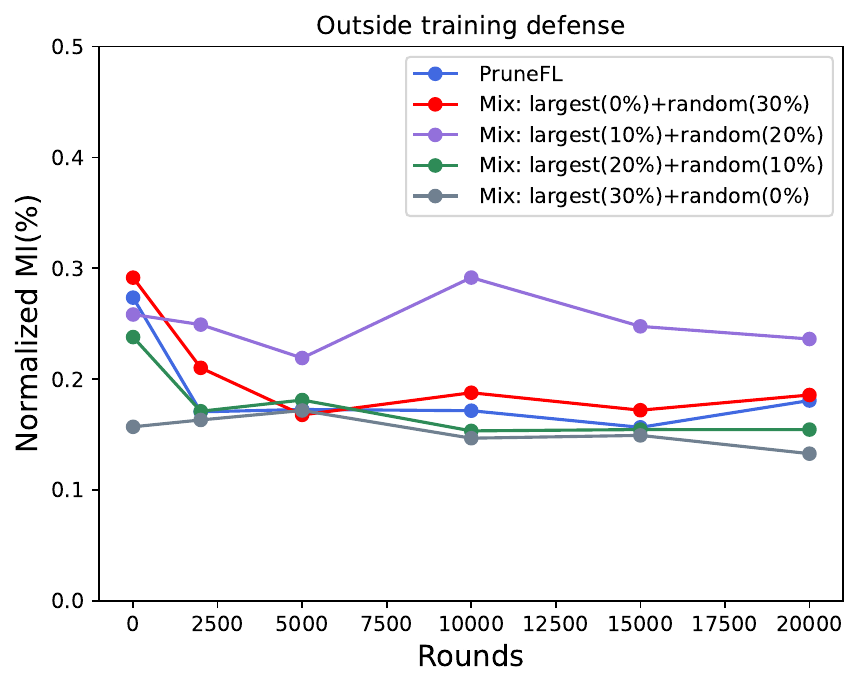}
    \caption{NMI Performance}
    \label{fig:nmi_mix}
  \end{subfigure}
  \hfill
  \begin{subfigure}[b]{0.3\textwidth}
    \includegraphics[width=\linewidth]{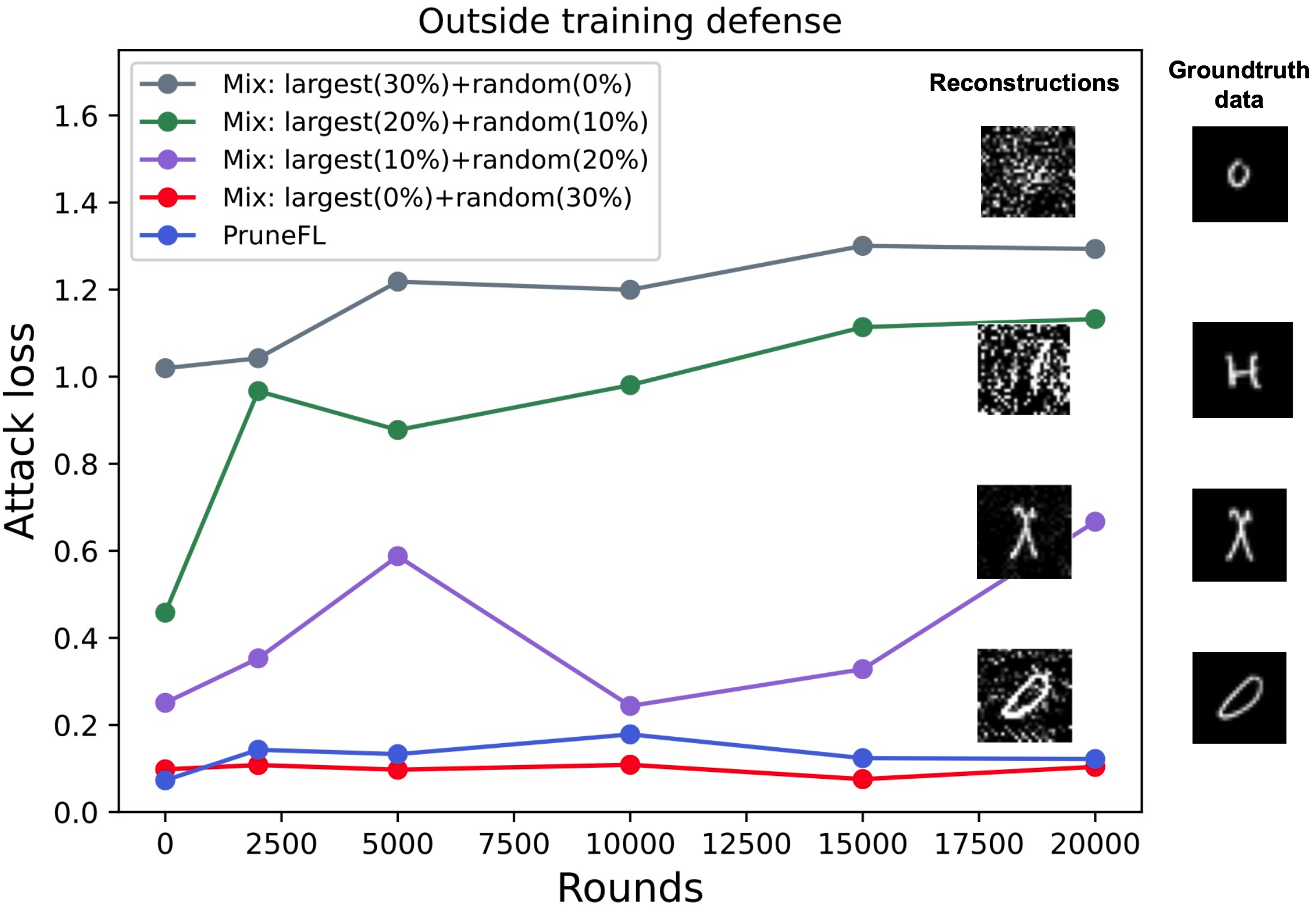}
    \caption{Attack Loss Performance}
    \label{fig:attack_loss_mix}
  \end{subfigure}
  \caption{Comparative Performance Analysis of Four Defense Rate Combinations in MIX Strategy across 20,000 FL Rounds.}
  \label{fig:mix_pruneFL}
\end{figure*}

Figure~\ref{fig:mix_pruneFL} shows the performance in terms of accuracy, privacy assessed through NMI, and attack loss of these four combinations across 20,000 iterations of FL. The outcomes reveal that the configuration of $0\%$ largest with $30\%$ random attains the highest accuracy, while $30\%$ largest with $0\%$ random yields the highest level of privacy protection. 

Since we focus on privacy in this work, we first proceed to investigate the privacy-accuracy trade-off for the $30\%$ largest with $0\%$ random  configurations.

\begin{figure}[!htbp]
  \centering
  \begin{subfigure}[b]{0.23\textwidth}
    \includegraphics[width=\linewidth]{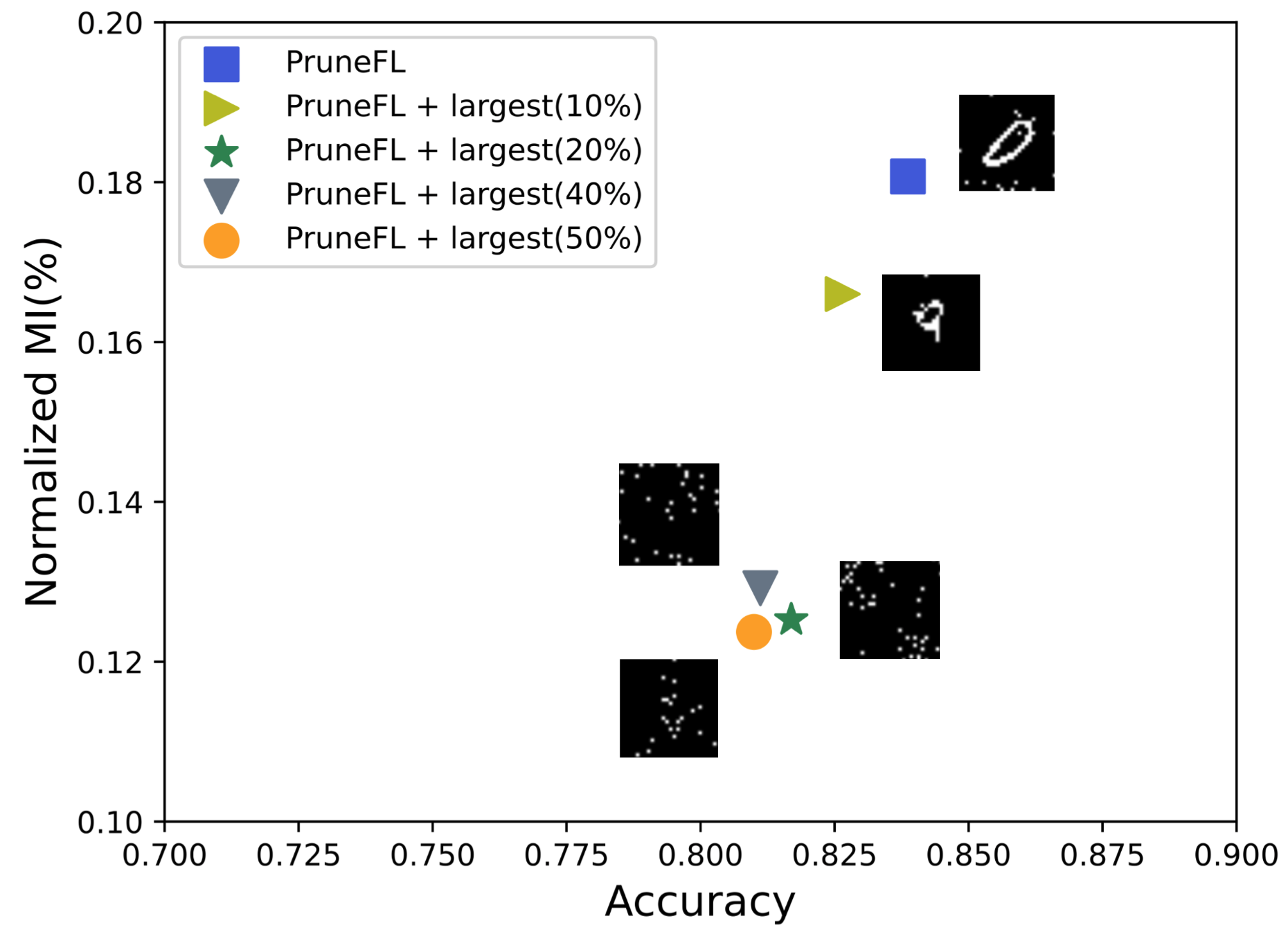}
    \caption{ACC vs NMI}
    \label{fig:subfig1}
  \end{subfigure}
  \hfill
  \begin{subfigure}[b]{0.23\textwidth}
    \includegraphics[width=\linewidth]{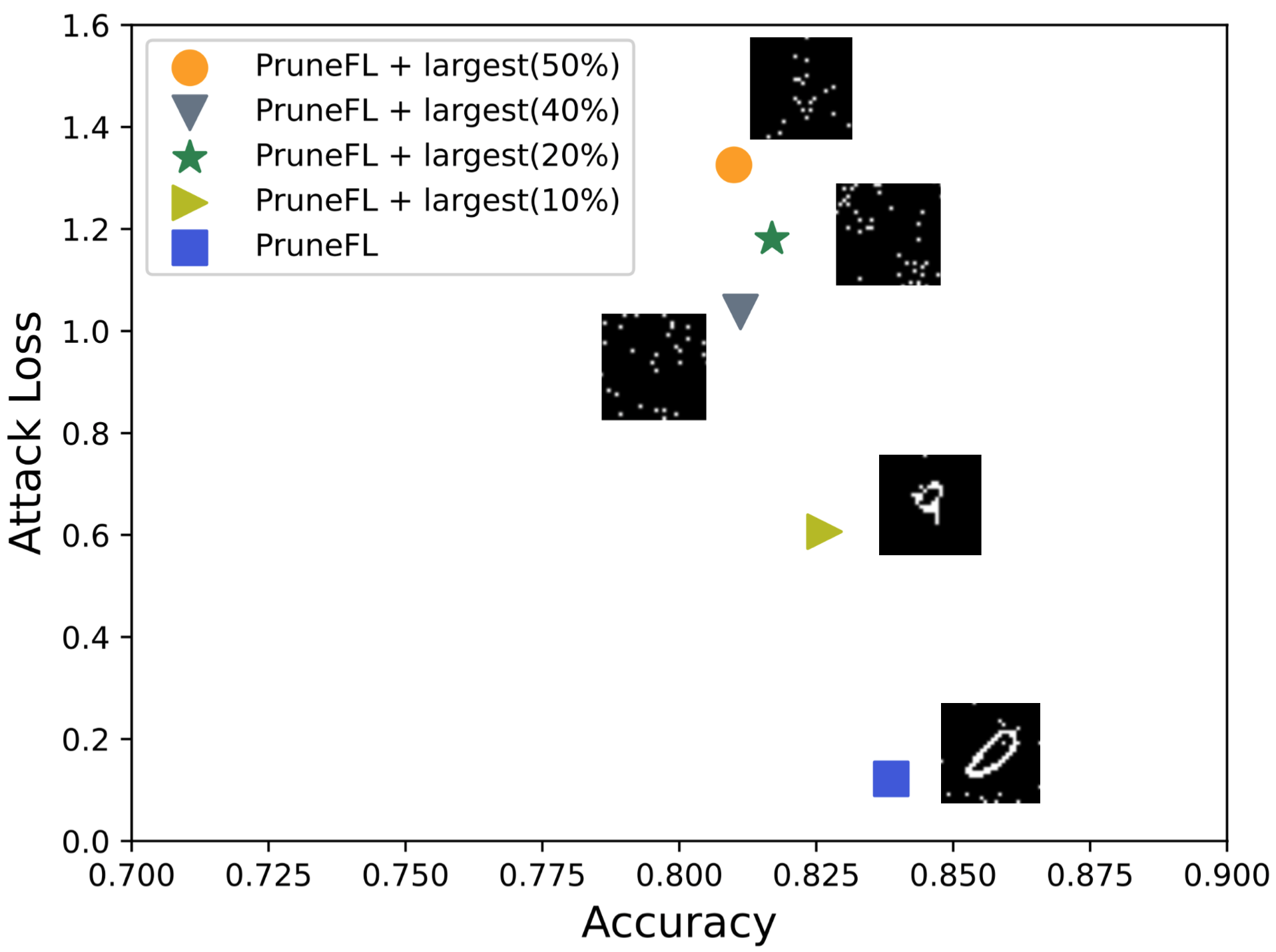}
    \caption{ACC vs Attack Loss}
    \label{fig:subfig2}
  \end{subfigure}
  
  \caption{Comparison of Privacy-Accuracy Trade-Off with Varying Defense Rate (0\% to 50\%) in Largest Method based on PruneFL using NMI and Attack Loss Privacy Metrics}
  \label{fig:trade_off_pruning_largest}
\end{figure}

Figure~\ref{fig:trade_off_pruning_largest} shows Privacy-Accuracy Trade-Off with Varying Defense Rate (0\% to 50\%) in Largest Method based on PruneFL using NMI and Attack Loss Privacy Metrics.

The outcomes are indicative of a notable trend: as more weights associated with larger gradients are pruned, privacy is enhanced at the expense of accuracy. This substantiates our \textbf{Insight 1}: Largest Weights Matter.
Pruning weights with large gradients improves privacy the most but also hurts model accuracy the most.

\begin{figure}[!htbp]
    \centering
    \centering
    \includegraphics[width=\linewidth]{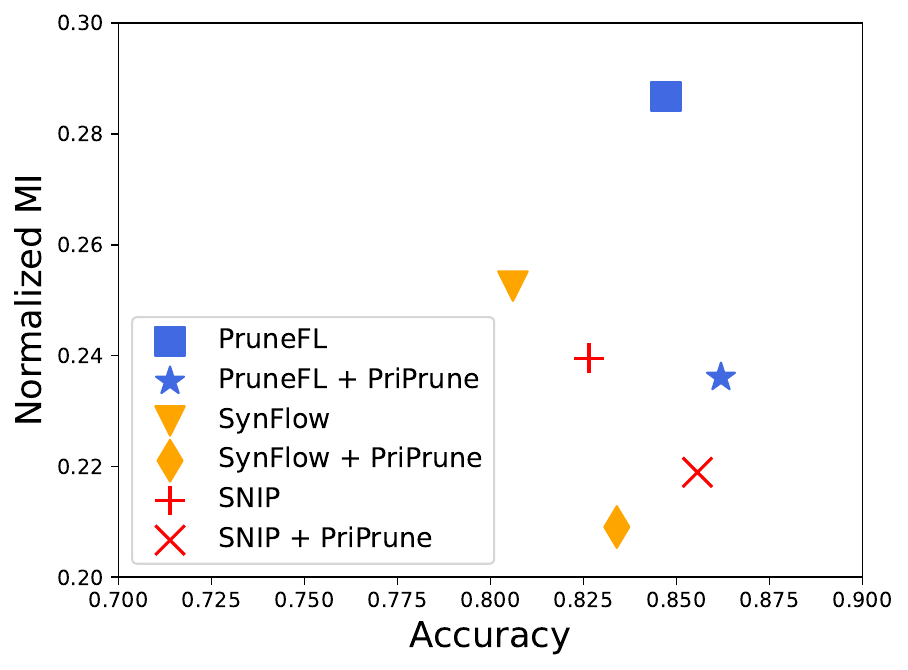}
    \caption{The whole trade-off analysis of integrating \algoname with various pruning mechanisms on CIFAR10 dataset}\label{fig:priprune_tradeoff_cifar}
\end{figure}



\subsubsection{\algoname's Performance on CIFAR10 Dataset} To show the effectiveness of \algoname, we also implement \algoname on CIFAR10 Dataset with various pruning methods as shown in Figure \ref{fig:priprune_tradeoff_cifar}. By incorporating \algoname, all pruning methods successfully elevate its privacy levels without undermining accuracy.

\end{document}